\documentclass[runningheads]{llncs}

\usepackage{amssymb}            % Defines common symbols like \mathbb R
\usepackage{mathtools}          % Extends amsmath, providing common math tools
\usepackage{mathrsfs}           % Enables \mathscr, which can work in cases that \mathcal does not
\usepackage{graphicx}           % For including images
\usepackage[space]{grffile}     % For spaces in image names
\usepackage[utf8]{inputenc} % allow utf-8 input
\usepackage[T1]{fontenc}    % use 8-bit T1 fonts
\usepackage{url}            % simple URL typesetting
\usepackage{booktabs}       % professional-quality tables
\usepackage{amsfonts}       % blackboard math symbols
\usepackage{nicefrac}       % compact symbols for 1/2, etc.
\usepackage{xcolor}         % colors
\usepackage{wrapfig}
\usepackage[sort&compress]{natbib}  % Other options: numbers, sort, compress
\usepackage{wrapfig}
\usepackage{subfigure}
\usepackage{caption}
\usepackage{enumitem}
\usepackage{amsmath}
\usepackage{tikz-cd}
\usepackage[colorlinks=true,citecolor=brown,linkcolor=blue,urlcolor=blue]{hyperref}
\usepackage{colortbl}
\usepackage{algorithm}
\usepackage[noend]{algorithmic}
\usepackage{bbding}
\usepackage{fancyvrb}
\usepackage{listings}
\usepackage[textsize=tiny, disable]{todonotes}
\definecolor{codegreen}{rgb}{0,0.5,0}
\definecolor{codered}{rgb}{0.7,0.1,0.1}
\definecolor{codegray}{rgb}{0.5,0.5,0.5}
\definecolor{codepurple}{rgb}{0.58,0,0.82}
\definecolor{backcolour}{rgb}{0.95,0.95,0.95}
\lstdefinestyle{python}{
    language=Python,
    basicstyle=\ttfamily\scriptsize,
    backgroundcolor=\color{backcolour},   
    commentstyle=\color{codegreen}\ttfamily\slshape,
    keywordstyle=\color{codered}\bfseries,
    numberstyle=\tiny\color{codegray},
    stringstyle=\color{codepurple},    
    xleftmargin=1em,
    framexleftmargin=0.3em,
    breakatwhitespace=false,         
    breaklines=true,                 
    captionpos=b,  % or t for top
    keepspaces=true,                 
    numbers=left,                    
    numbersep=4pt,                  
    showspaces=false,                
    showstringspaces=false,
    showtabs=true,                  
    tabsize=1,
    fancyvrb=true
}
\newcounter{todocounter}

\newcommand{\todoil}[1]{\todo[inline, size=\small, color=orange!50]{#1}}
\newcommand{\todozlf}[2][]{\todo[color=yellow!80!black, #1]{LF: #2}}
\newcommand{\todoilzlf}[2][]{\todo[inline, size=\small, color=yellow!80!black, #1]{LF: #2}}
\newcommand{\todonote}[1]{\todo[color=blue!50!white]{#1}}

\newcommand{\todolsw}[1]{\todo[color=green!80!black]{LSW: #1}}

\newcommand{\edit}[1]{{\color{black} #1}}
\newcommand{\ethree}{{\mathrm{E}(3)}}

\newcommand{\etwo}{{\mathrm{E}(2)}}
\newcommand{\setwo}{{\mathrm{SE}(2)}}
\newcommand{\sothree}{{\mathrm{SO}(3)}}
\newcommand{\sotwo}{{\mathrm{SO}(2)}}
\newcommand{\otwo}{{\mathrm{O}(2)}}
\newcommand{\ztwo}{{\mathbb{Z}^2}}
\newcommand{\rtwo}{{\mathbb{R}^2}}
\newcommand{\rthree}{{\mathbb{R}^3}}
\DeclareMathOperator{\Hom}{Hom}

\usepackage{amsmath,amsfonts,bm}

\def\eqref#1{equation~\ref{#1}}
\def\1{\bm{1}}

\def\va{{\bm{a}}}

\def\vs{{\bm{s}}}

\def\vz{{\bm{z}}}

\DeclareMathAlphabet{\mathsfit}{\encodingdefault}{\sfdefault}{m}{sl}
\SetMathAlphabet{\mathsfit}{bold}{\encodingdefault}{\sfdefault}{bx}{n}

\def\gA{{\mathcal{A}}}

\def\gE{{\mathcal{E}}}

\def\gS{{\mathcal{S}}}
\def\gT{{\mathcal{T}}}

\def\gV{{\mathcal{V}}}

\def\sA{{\mathbb{A}}}

\def\sR{{\mathbb{R}}}

\def\sZ{{\mathbb{Z}}}

\DeclareMathOperator*{\argmax}{arg\,max}
\DeclareMathOperator*{\argmin}{arg\,min}

\begin{document}

% \title{Equivariant Sampling Methods in Reinforcement\\ Learning and Planning}
% \title{Euclidean Equivariant Action Sampling for\\ Reinforcement Learning and Planning}
\title{Equivariant Action Sampling for\\ Reinforcement Learning and Planning}


\author{
Linfeng Zhao\inst{1} \and
Owen Howell\inst{1} \and
Xupeng Zhu\inst{1} \and
Jung Yeon Park\inst{1} \and
Zhewen Zhang\inst{1} \and
Robin Walters$^\dagger$\inst{1} \and
Lawson L.S. Wong$^\dagger$\inst{1}
}

\authorrunning{L. Zhao et al.}
\institute{
Northeastern University, Boston, MA, USA\\
\email{zhao.linf@northeastern.edu}, \email{robin.walters@northeastern.edu}, \email{lsw@ccs.neu.edu} %\\
% \url{http://lfzhao.com/SymCtrl}
% \url{http://lfzhao.com/ESC}
}

%%%%%%%%%%%%%%%%%%%%%%%%%%%%%%%%%%%%%%%%%%%%%%%%%%%%%%%%%%%%%%%%
%% Begin document, create title and abstract
%%%%%%%%%%%%%%%%%%%%%%%%%%%%%%%%%%%%%%%%%%%%%%%%%%%%%%%%%%%%%%%%
% \begin{document}

\maketitle

\begin{abstract}

Reinforcement learning (RL) algorithms for continuous control tasks require accurate sampling-based action selection. Many tasks, such as robotic manipulation, contain inherent problem symmetries. However, correctly incorporating symmetry into sampling-based approaches remains a challenge. This work addresses the challenge of preserving symmetry in sampling-based planning and control, a key component for enhancing decision-making efficiency in RL. We introduce an action sampling approach that enforces the desired symmetry. We apply our proposed method to a coordinate regression problem and show that the symmetry aware sampling method drastically outperforms the naive sampling approach. We furthermore develop a general framework for sampling-based model-based planning with Model Predictive Path Integral (MPPI). We compare our MPPI approach with standard sampling methods on several continuous control tasks. Empirical demonstrations across multiple continuous control environments validate the effectiveness of our approach, showcasing the importance of symmetry preservation in sampling-based action selection.


\end{abstract}

\addtocontents{toc}{\protect\setcounter{tocdepth}{-1}}

%%%%%%%%%%%%%%%%%%%%%%%%%%%%%%%%%%%%%%%%%%%%%%%%%%%%%%%%%%%%%%%%%%%%%%%%%%%%%%%%%%%%%%%%%
%%%%%%%%%%%%%%%%%%%%%%%%%%%%%%%%%%%%%%%%%%%%%%%%%%%%%%%%%%%%%%%%%%%%%%%%%%%%%%%%%%%%%%%%%

\section{Introduction}

In reinforcement learning (RL) for continuous control, the need for effective sampling-based action selection is paramount. Many control environments, especially in robotic manipulation and navigation, exhibit symmetries due to their operation within Euclidean space. While previous explorations of equivariance have primarily focused on deterministic RL policies \citep{ravindran2004algebraic,zinkevich_symmetry_2001,van2020mdp,mondal_group_2020,wang_mathrmso2-equivariant_2021,zhao_integrating_2022}, the inherently multimodal nature of these control tasks demands sampling-based approaches. This crucial integration of symmetry into sampling-based methods remains largely under-explored.

In general, sampling methods will break the exact symmetries of action selection. This is an issue as the breaking of symmetry prevents the use of equivariant reinforcment learning methods \citep{van_der_pol_mdp_2020}.
% In this work, we show how sampling methods can be designed to enforce end-to-end equivariance, leading to improved sample efficiency.
%
This paper addresses the challenge of preserving symmetry in sampling-based planning and control, a vital aspect for enhancing decision-making efficiency in RL. Specifically, existing sampling methods maintain symmetry only in the limit of an infinite number of samples. In the case of finite samples, problem symmetries are only conserved approximately.

%failing to do so in the more realistic scenarios of finite or individual samples. 
%
We formulate the action optimization challenge as a two-step procedure: first, we estimate the performance of state-action pairs or trajectories using a neural network like Q-values, energy functions, or returns; then, we engage in sampling to optimize optimal actions or action sequences. This paper investigates the equivariance properties of this formulation, crucial for the effectiveness of sampling-based strategies in symmetric environments.

% Our exploration begins with coordinate regression, a simplified problem that provides foundational insights into equivariant action sampling. Through this lens, we visualize prediction and energy functions, offering a concrete demonstration of our methodology's potential.

% Extending our insights to Model Predictive Path Integral (MPPI) for multi-step action selection presents a natural progression. MPPI, which utilizes n-step returns to evaluate trajectories, necessitates a nuanced understanding of symmetry preservation across action sequences, dynamics, reward, and policy functions. This extension not only challenges but also enriches our approach to maintaining equivariance in more complex scenarios.

In this study, we first study the coordinate regression problem, providing profound insights into the mechanisms of equivariant action sampling. 
% Through this problem, we visualize the prediction and energy functions, providing a tangible illustration of how symmetry impacts performance.
We find that using a symmetry invariant energy function alongside our novel sampling strategy significantly enhances algorithm performance. Our strategy deviates from conventional sampling methods by augmenting sampled actions with the symmetry group $G$. This ensures that the sampling procedure always preserves equivariance, irrespective of the number of samples. Without our proposed sampling strategy, the symmetry of the two-step procedure holds only in the infinite sample limit.
We extend our sampling strategy to multi-step action selection for sampling-based planning.
We propose an equivariant version of Model Predictive Path Integral (MPPI)~\citep{williams_information_2017} and derive that it needs equivariant dynamics and reward model, and equivariant policy and value network, analogous to the need of equivariant energy function.
Based on it, we implement an equivariant model-based RL algorithm, TDMPC \citep{hansen_temporal_2022}, for continuous control tasks.
This adaptation allows for a comprehensive preservation of symmetry across the entire trajectory planning process. 
% Our approach not only demonstrates the feasibility of integrating equivariance into MPPI but also highlights the significant advantages this integration offers for enhancing the efficiency and effectiveness of model-based reinforcement learning in symmetric continuous control environments.

The contributions of this paper include offering both theoretical insights into equivariance in sampling-based planning and practical demonstrations of a novel equivariant sampling methodology's effectiveness. Empirical validations across various continuous control environments underscore the significance of our findings, illuminating the path for future research in symmetric RL tasks.

%%%%%%%%%%%%%%%%%%%%%%%%%%%%%%%%%%%%%%%%%%%%%%%%%%%%%%%%%%%%%%%%%%%%%%%%%%%%%%%%%%%%%%%%%%%%%%%%
%%%%%%%%%%%%%%%%%%%%%%%%%%%%%%%%%%%%%%%%%%%%%%%%%%%%%%%%%%%%%%%%%%%%%%%%%%%%%%%%%%%%%%%%%%%%%%%%
\section{Related Work}

% Change logs:
% WAFR - make as a separate section and add more related work originally from appendix
% made the section more coherent

\paragraph{Symmetry in Reinforcement Learning}
Symmetry in decision-making tasks has been studied in the context of reinforcement learning and control. Early research focused on symmetry in MDPs without function approximation \citep{ravindran2004algebraic,ravindran_symmetries_nodate,zinkevich_symmetry_2001}, while more recent work has explored symmetry in model-free (deep) RL and imitation learning using equivariant policy networks \citep{van2020mdp,mondal_group_2020,wang_mathrmso2-equivariant_2021,Huang_2024,wang2022onrobotlearningequivariantmodels,xie2020deep,Jia2024openvocabulary,sortur2023sample,zhao2023euclidean}. 
\citet{park_learning_2022} investigated equivariance in learning world models.
Additionally, \citet{zhao_integrating_2022} analyzed the use of symmetry in value-based planning on a 2D grid. Our work extends these studies by focusing on continuous-action MDPs and sampling-based planning algorithms.

\paragraph{Geometric Graphs and Geometric Deep Learning}
Our definition of GMDP is closely related to the concept of \textit{geometric graphs} \citep{bronstein_geometric_2021,brandstetter_geometric_2021}, which model MDPs as state-action connectivity graphs and have been used to examine algorithmic alignments and dynamic programming \citep{xu_what_2019,dudzik_graph_2022}. We extend this concept by embedding MDPs into geometric spaces such as $\rtwo$ or $\rthree$, focusing on 2D and 3D Euclidean symmetry \citep{lang_wigner-eckart_2020,brandstetter_geometric_2021,weiler_general_2021} with corresponding symmetry groups $\etwo$ and $\ethree$. Geometric deep learning, which maintains geometric properties like symmetry and curvature in data analysis \citep{bronstein_geometric_2021}, has developed equivariant neural networks to preserve these symmetries. Notable contributions include G-CNNs \citep{cohen_group_2016}, which introduced group convolutions, and Steerable CNNs \citep{cohen_steerable_2016}, which generalize scalar feature fields to vector fields and induced representations. Additionally, $E(2)$-CNNs \citep{weiler_general_2021} solve kernel constraints for $E(2)$ and its subgroups by decomposing into irreducible representations. Researchers have also explored steerable message-passing GNNs \citep{brandstetter_geometric_2022}, $E(n)$-equivariant graph networks \citep{satorras_en_2021}, and the theory of equivariant maps and convolutions for scalar and vector fields \citep{kondor_generalization_2018,cohen_general_2020}. Recent work has focused on designing latent equivariant architectures for 3D scene renderings \citep{Klee_2023,Howell_2023}.

\paragraph{Planning in Reinforcement Learning}
Planning in RL involves devising strategies to achieve long-term goals by predicting future states and rewards. 
MuZero~\citep*{schrittwieser_mastering_2019} uses 
TDMPC~\citep{hansen_temporal_2022} integrates planning using Model Predictive Path Integral~\citep{williams_model_2017} with temporal-difference learning~\citep{sutton_reinforcement_2018}.
Planning has also played vital roles in robotics, where high-level task planning and geometric-level motion planning are crucial for enabling robots to perform complex tasks autonomously \citep{garrett_integrated_2020,kumar2024practice,zhao2024learning}.
These planning techniques are crucial for enabling agents and robots to perform complex tasks autonomously. However, they typically do not consider symmetry, which can lead to inefficiencies and suboptimal performance in symmetric environments.

%%%%%%%%%%%%%%%%%%%%%%%%%%%%%%%%%%%%%%%%%%%%%%%%%%%%%%%%%%%%%%%%%%%%%%%%%%%%%%%%%%%%%%%%%
%%%%%%%%%%%%%%%%%%%%%%%%%%%%%%%%%%%%%%%%%%%%%%%%%%%%%%%%%%%%%%%%%%%%%%%%%%%%%%%%%%%%%%%%%

\section{MDPs with Geometric Structure}
\label{sec:statement-gmdp}

% NOTE changelogs:
% removed the geometric MDP figure
% removed the geometric MDP examples, explanation - moved all explanation to appendix; add reference here + briefly explain table
% removed additional theory section; merge remained stuff with formulation sec
% removed entire linearized MDP theory part, after current Sec 2.1
% removed theory section; merge remained stuff with formulation sec

\subsection{Formulation of Geometric MDPs}

% \todonote{consider rewrite the subsection: highlight \textbf{latent} geometric structure}

Markov Decision Processes (MDPs) are fundamental in modeling decision-making in interactive environments. In robotic control applications, MDPs often involve state spaces defined over Euclidean spaces $\mathbb{R}^d$ or on groups like $\mathrm{SE}(d)$. These state spaces might directly represent physical spaces, such as the position of a robot, or embody latent structures in sensor inputs, such as camera images.

The study of isometric changes, which are transformations that preserve distances in the state space, introduces the Euclidean symmetry group $\mathrm{E}(d)$. This group and its subgroups, which can be expressed in semi-direct product form as $\left(\mathbb{R}^d,+\right) \rtimes G$, play a crucial role in how we understand and manipulate these state spaces. The group action, consisting of translations and rotations or reflections, transforms a vector $x$ in the space according to $x \mapsto (t g) \cdot x := g x + t$ \citep{weiler_general_2021,lang_wigner-eckart_2020}. These transformations are critical for defining symmetry properties in the system, which lead to more efficient problem-solving strategies~\citep{van_der_pol_mdp_2020,wang_mathrmso2-equivariant_2021,zhao_integrating_2022,teng_error-state_2023}.

We define a class of MDPs with internal geometric structure, where the ground state space or a latent space of the MDP can be transformed by a Euclidean group. This extends a previously studied discrete case \citep{zhao_integrating_2022}.

\begin{definition}[Geometric MDP] \label{def:gmdp}
A Geometric MDP (GMDP) $\mathcal{M}$ is an MDP with internal geometric structure: there is a symmetry group $G \leq \operatorname{GL}(d)$ that acts on the ground or latent state space $\gS$ and action space $\gA$. It is written as a tuple $\langle \gS, \gA, P, R, \gamma, G, \rho_\mathcal{S}, \rho_\mathcal{A} \rangle$. The state and action spaces $\gS, \gA$ have group actions that transform them, defined by $\rho_\gS$ and $\rho_\gA$.
\end{definition}

% NOTE: we removed anything related to continuous group action! also remove compact group!
% NOTE: previous version:
% A Geometric MDP (GMDP) $\mathcal{M}$ is an MDP that the state and action features are vectors in the Euclidean space $\sR^d$ and are equipped with a (compact) symmetry group $G \leq \operatorname{GL}(d)$ that acts on them.
% A Geometric MDP (GMDP) $\mathcal{M}$ is an MDP that is embedded into a Euclidean? space and is equipped with a (compact) symmetry group $G \leq \operatorname{GL}(d)$ that acts on the space.
% It is written as a tuple $\langle \gS, \gA, P, R, \gamma, \gB, G, \rho_\mathcal{S}, \rho_\mathcal{A} \rangle$.
% The state and action spaces $\gS, \gA$ have (continuous) group actions that transform them, defined by $\rho_\gS$ and $\rho_\gA$.
% There exists a map $\gS \times \gA \mapsto \gB$ that "squeeze" the state-action space to a smaller base space $\gB$.

% \todoil{explain the definition a bit more?}

\subsection{Symmetry in Geometric MDPs}

% CHANGELOG
% > make a separate subsection here, state our theorems here

In this subsection, we explore how Euclidean symmetry influences the internal geometric structure of MDPs.
The symmetry properties in MDPs are characterized by the equivariance and invariance of the transition and reward functions, respectively \citep{zinkevich_symmetry_2001,ravindran2004algebraic,van2020mdp,wang_mathrmso2-equivariant_2021,zhao_integrating_2022,zhao2024e2plan}:
\begin{align} \label{eq:symmetry-mdp}
&\forall g \in G, \forall s, a, s',  &{P}(s' \mid s, a )  &= {P}(g \cdot s' \mid g \cdot s, g \cdot a )  \\
&\forall g \in G, \forall s, a, &{R}(s, a ) &= {R}(g \cdot s, g \cdot a )  
\end{align}

Here, $g$ acts on the state and action spaces through group representations $\rho_\gS$ and $\rho_\gA$, respectively.
For instance, the standard representation $\rho_\text{std}(g)$ of $\sotwo$ assigns each rotation $g \in \sotwo$ a 2D rotation matrix $R_{2 \times 2}(g)$ given by:
\begin{align*}
R(g) = \begin{bmatrix}
\cos(g) & \sin(g) \\
-\sin(g) & \cos(g) 
\end{bmatrix}
\end{align*}
This matrix represents a rotation by an angle $g$.
The trivial representation $\rho_\text{tri}(g)$ assigns the one-dimensional identity matrix $\rho_{tri}(g) = \mathbf{1}_{1 \times 1}$ to all $g$.

\begin{figure}[t]
\centering
\includegraphics[width=1.0\linewidth]{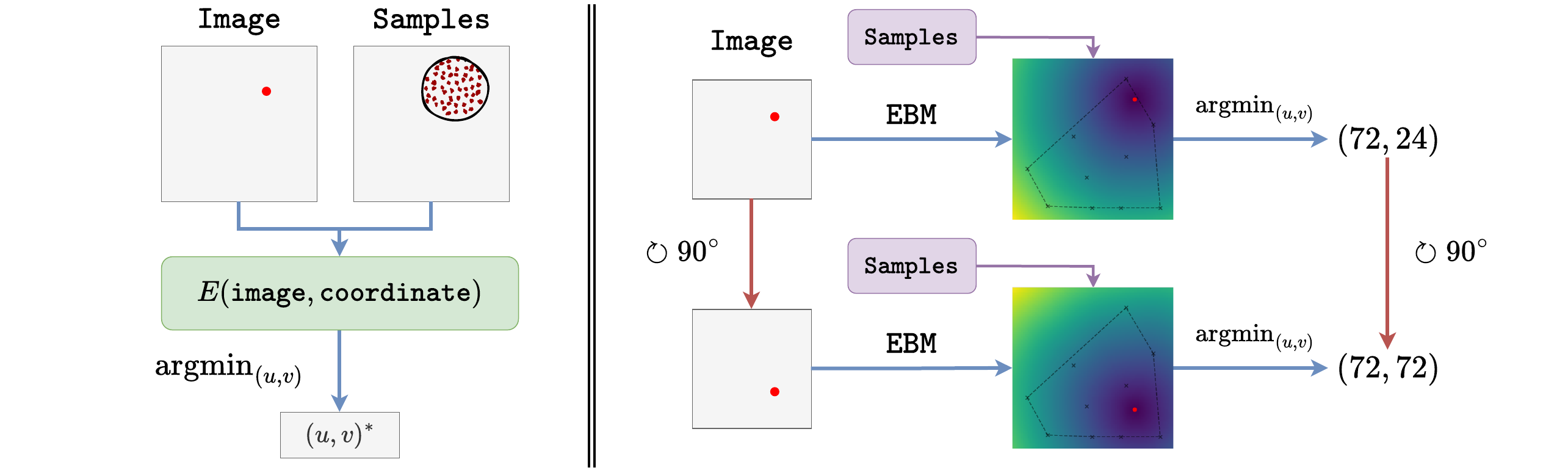}
\caption{
\small
Illustration of the coordinate regression problem (Sec~\ref{subsec:def-coord-regression}) and its equivariance.
(Left) The energy function \texttt{EBM} takes \textbf{image} and coordinate \textbf{samples} and outputs scalar energy value.
(Right) Equivariance in coordinate regression: rotating the image and augmenting samples results in rotated coordinate prediction.
}
\vspace{-5pt}
\label{fig:algo-equiv-sampling}
\end{figure}

\paragraph{Properties.}
In a geometric MDP with a discrete symmetry group, the optimal policy mapping is $G$-equivariant, as demonstrated in \citep{ravindran2004algebraic}.
To incorporate symmetry constraints, one strategy is to ensure the entire policy mapping is equivariant: $a_t = \texttt{policy}(s_t)$ \citep{van_der_pol_mdp_2020,wang_mathrmso2-equivariant_2021,zhao_integrating_2022,zhao2024e2plan}, as illustrated in Figure~\ref{fig:algo-equiv-sampling}.

Many model-based RL algorithms rely on iteratively applying Bellman operations \citep{sutton_reinforcement_2018}.
We show that the symmetry $G$ in a Geometric MDP (GMDP) results in a $G$-equivariant Bellman operator.
This implies that the iterative process in model-based RL algorithms can be constrained to be $G$-equivariant to exploit symmetry.

Additionally, for GMDPs, a specific instance of a dynamic programming (DP)-based algorithm, value iteration, can be connected with geometric graph neural networks \citep{bronstein_geometric_2021}.
For non-geometric graphs, \citet{dudzik_graph_2022} demonstrated the equivalence between dynamic programming on a general non-geometric MDP and a message-passing GNN.

\begin{theorem}
The Bellman operator of a geometric MDP is equivariant under the Euclidean group $\mathrm{E}(d)$, which includes $d$-dimensional isometric transformations.
\end{theorem}

We provide proofs and derivations in Appendix~\ref{sec:theory-appendix}.
This extends the theorems in \citep{zhao_integrating_2022} on 2D discrete groups, where they showed that value iteration is equivariant under discrete subgroups of the Euclidean group, such as discrete translations, rotations, and reflections.
We generalize this result to groups of the form $\left(\sR^d,+\right) \rtimes G$, where $G$ includes continuous rotations and translations.\footnote{For the translation part, one may use relative/normalized positions or induced representations \citep{cohen_general_2020,lang_wigner-eckart_2020}.}

\subsection{Illustration and Examples of Geometric MDPs}
\label{subsec:appendix-examples-g-mdps}

\begin{table}[t]
\centering
\caption{
\small
Examples of tasks modeled as geometric MDPs.
$G$ denotes the MDP symmetry group, $\mathcal{S}$ denotes the MDP state space, and $\mathcal{A}$ denotes the MDP action space.
We can quantitatively measure the savings of equivariance.
"Images" refers to panoramic egocentric images $\ztwo \to \sR^{H \times W \times 3}$.
$\circ$ denotes group element composition.
We list the quotient space $\gS/G$ to provide intuition on savings.
The $Gx = \{ g \cdot x \mid g \in G \}$ column shows the $G$-orbit space of $\gS$ ($\cong$ denotes isomorphic to).
}
\scriptsize
\begin{tabular}{@{}llllllllll@{}}
\toprule
ID & $G$        & $\gS$                                       & $\gA$      & $\gS / G$                 & $Gx$ & Task                                                   \\ \midrule
1 & $C_4$      & $\ztwo$                                     & $C_4$      & $\ztwo/C_4$          & $C_4$        & 2D Path Planning \citep{tamar_value_2016} \\
2 & $C_4$      & Images                                      & $C_4$      & $\ztwo/C_4$          & $C_4$        & 2D Visual Navigation  \citep{zhao_integrating_2022}                                     \\ \midrule
3 & $\sotwo$   & $\rtwo$                                     & $\rtwo$    & $\sR^+$              & $S^1$    & 2D Continuous Navigation \citep{zhao2024e2plan}                                           \\
4 & $\sothree$ & $\rthree \times \rthree$                    & $\rthree$  & $\sR^+ \times \sR^3$ & $S^2$    & 3D Free Particle (with velocity)                                                 \\
5 & $\sothree$ & $\rthree \rtimes \sothree$                  & $\rthree \times \rthree$  & $\sR^+ \times \sR^3$ & $S^2$    & Moving 3D Rigid Body                                                 \\
6 & $\sotwo$   & $\sotwo$                                    & $\rtwo$    & \textbf{$\{ e \}$}   & $S^1$    & Free Particle on $\sotwo \cong S^1$ manifold                               \\
7 & $\sothree$ & $\sothree$                                  & $\rthree$  & \textbf{$\{ e \}$}   & $S^2$    & Free Particle on $\sothree$ \citep{teng_error-state_2023}                              \\
8 & $\sotwo$   & $\setwo$                                    & $\setwo$   & $\sR^2$              & $S^1$    & Top-down Grasping   \citep{zhu_sample_2022}                         \\ 
9 & $\sotwo$   & $(S^1)^2 \times (\rtwo)^2$                  & $\sR^2$    & $S^1 \times (\rtwo)^2$ & $S^1$    & Two-arm Manipulation  \citep{tassa_deepmind_2018}    \\ 
\bottomrule
\end{tabular}
\label{tab:appendix-examples-geometric-mdps}
\vspace{-10pt}
\end{table}

Geometric MDP examples include moving a point robot in a 2D continuous space ($\sR^2$, Example 3 in Table~\ref{tab:appendix-examples-geometric-mdps}) or a discrete space ($\sZ^2$, Example 1 \citep{tamar_value_2016}), which is the abstraction of 2D discrete or continuous navigation.
Table~\ref{tab:appendix-examples-geometric-mdps} includes more relevant examples.
We use visual navigation over a 2D grid ($\sZ^2 \rtimes C_4$, Example 2 \citep{lee_gated_2018, zhao_integrating_2022}) as another example of a Geometric MDP.
In this example, each position in $\sZ^2$ and orientation in $C_4$ has an image in $\sR^{H \times W \times 3}$, which is a \textit{feature map} $\sZ^2 \rtimes C_4 \to \sR^{H \times W \times 3}$.
The agent only navigates on the 2D grid $\sZ^2$ (potentially with an orientation of $C_4$), but not the raw pixel space.
Example (4) extends to the continuous 3D space and also includes linear velocity $\rthree$.
Alternatively, we can consider (5) moving a rigid body with $\sothree$ rotation.
In (6) and (7), we consider moving free particle positions on $\sotwo,\sothree$, which are examples of optimal control \textit{on manifold} in \citep{lu_-manifold_2023,teng_error-state_2023}.
Here, $G=\sothree$ acts on $\gS=\sothree$ by group composition.
(8) top-down grasping needs to predict $\setwo$ action on grasping an object on a plane with $\setwo$ pose.
It additionally has translation symmetry, so the state space is technically $\setwo/\setwo=\{ e \}$.
(9) is the \texttt{Reacher} task, where the agent controls a two-joint arm.
It is easy to see that because two links are connected, kinematic constraints restrict the potential possible improvement with an equivariant algorithm.
Additionally, Example (3) is later implemented as \texttt{PointMass}, which is (8) top-down grasping without $\sotwo$ rotation.

%%%%%%%%%%%%%%%%%%%%%%%%%%%%%%%%%%%%
% NOTE: Originally we have equivariance in planning figure here

%%%%%%%%%%%%%%%%%%%%%%%%%%%%%%%%%%%%

%%%%%%%%%%%%%%%%%%%%%%%%%%%%%%%%%%%%%%%%%%%%%%%%%%%%%%%%%%%%%%%%%%%%%%%%%%%%%%%%%%%%%%%%%%%%%%%%%%%%%%
%%%%%%%%%%%%%%%%%%%%%%%%%%%%%%%%%%%%%%%%%%%%%%%%%%%%%%%%%%%%%%%%%%%%%%%%%%%%%%%%%%%%%%%%%%%%%%%%%%%%%%
\section{Equivariant Sampling-Based Action Selection}
\label{sec:equiv-action-sampling}

\subsection{Action Selection via Sampling}
\label{subsec:action-sampling-def}

Many real-world reinforcement learning (RL) and imitation learning problems involve continuous actions $\va_t \sim \pi ( a \mid \vs_t)$. When the action space $\gA$ is infinite, the policy function may need to employ stochastic sampling.

Concretely, we focus on a two-step ``implicit policy'' strategy~\citep{kalashnikov_qt-opt_2018,florence_implicit_2021,hansen_temporal_2022}, illustrated in Figure~\ref{fig:algo-equiv-sampling} (left), which solves the action optimization problem $a^* = \argmin_{a} E(s,a)$ via sampling in two steps:
(1) A neural network evaluates state-action pairs $(\vs_t, \va_t)$ or trajectories $(\vs_t, \va_t, \vs_{t+1}, \va_{t+1}, \ldots)$;
(2) Online optimization is performed over the action space $\va \in \gA$ to select better actions for each state $\vs_t$ using methods such as the Cross-Entropy Method (CEM) \citep{de_boer_tutorial_2005} or Model-Predictive Path Integral \citep{williams_model_2017}. In step (1), the neural network can be parameterized as an energy function $E_\theta(s,a)$ being minimized (Implicit Behavior Cloning, IBC; \citet{florence_implicit_2021}), a $Q$-value network $Q_\theta(s,a)$ being maximized (QT-Opt; \citet{kalashnikov_qt-opt_2018}), or the $N$-step return of a trajectory being maximized (TD-MPC; \citet{hansen_temporal_2022}).

\subsection{Coordinate Regression Problem}
\label{subsec:def-coord-regression}

In learning visuo-motor policies, a key challenge is converting high-dimensional image data into continuous action outputs. This challenge is exemplified in the coordinate regression problem \citep{florence_implicit_2021}, where the goal is to predict the \( xy \) coordinates of a specific target marker within an image, as shown in Figure~\ref{fig:algo-equiv-sampling} (left). We use this problem to demonstrate and study the equivariance properties of action sampling.

In the coordinate regression problem, the objective is to predict the $(x,y)$ coordinate value $v$ of a (green) marker on an image $I$: $v^* = \argmax_{v \in \sR^2} E_\theta(I, v)$, which can be written as a function $h(\cdot)$ of the image $v^* = h(I)$.

The coordinate regression problem exhibits both rotation and reflection symmetry (or $g \in D_4$ dihedral group). Specifically, if the input image is rotated/flipped $g \cdot I$, the network should predict the rotated/flipped coordinate value $g \cdot v$: $h_\theta(g \cdot I) = g \cdot h_\theta(I) \equiv g \cdot v$.
Although simplified, the coordinate regression problem captures many fundamental challenges inherent in visuomotor control problems like robotic manipulation and control. Understanding the interplay between symmetry and sampling in this problem will help build intuition for equivariant action sampling in real-world tasks.

%%%%%%%%%%%%%%%%%%%%%%%%%%%%%%%%%%%%
\begin{figure*}[t]
\centering
%\vspace{-20pt}
\subfigure{
\includegraphics[width=1.\linewidth]{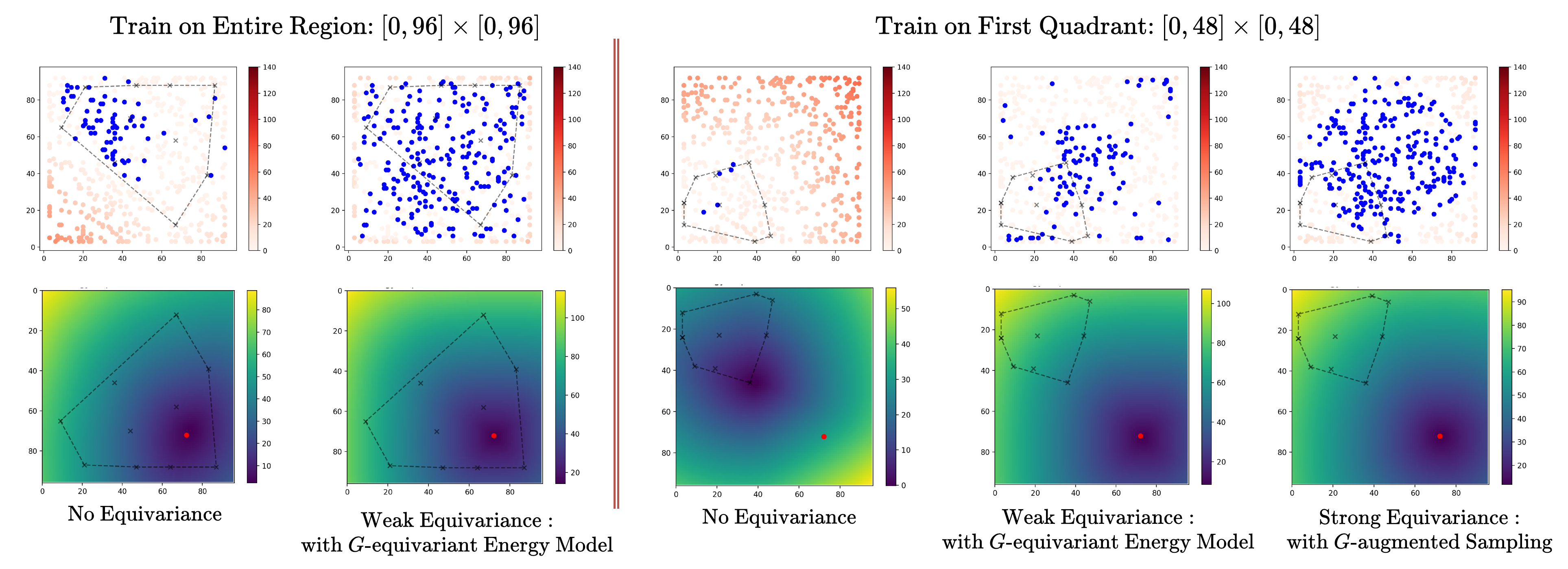}
}
\vspace{-15pt}
\centering
\caption{
\small
Demonstration of results on coordinate regression problem: left two columns for training on entire region, and right three columns for training only on coordinates in first quadrant.
}
\vspace{-15pt}
\label{fig:results-coordinate-regression}
\end{figure*}
%%%%%%%%%%%%%%%%%%%%%%%%%%%%%%%%%%%%

\subsection{Strong and Weak Equivariance in Sampling}

We can classify sampling methods that satisfy symmetry constraints as either \textbf{weak equivariance} or \textbf{strong equivariance}. Suppose that we are trying to estimate the function $f(x)$ via sampling 
\begin{align*}
f(x) = \mathbb{E}_{\omega}[  q(x, \omega)  ],
\end{align*}
where $\omega$ is drawn from some probability distribution and $q(x,\omega)$ averaged over the distribution of $\omega$ returns $f(x)$. Now, suppose that we know a-priori that the function $f$ satisfies some equivariant constraints,
\begin{align*}
\forall g \in G, \quad f( g \cdot x ) = \rho(g^{-1}) f(x).
\end{align*}
This constraint arises naturally in many energy based models, where the energy function is invariant under some spatial symmetry, see Fig.~\ref{fig:algo-equiv-sampling}. In the naive approach, we can drawn $m$ sample points $\omega_{1},\omega_{2}, ... , \omega_{m}$ i.i.d. and estimate the sample average $\hat{f}(x)$ of the function $f(x)$ as
\begin{align*}
\hat{f}(x) = \frac{1}{m} \sum_{i=1}^{m} q( x , \omega_{i} ).
\end{align*}
However, this approximation $\hat{f}(x)$ does not need to satisfy the original equivariance property. We will say that a sample estimator $\hat{f}$ of a $G$-equivariant function $f$ is \textbf{weakly equivariant} if 
\begin{align}\label{weak-estimator}
\forall g \in G, \quad \mathbb{E}_{\omega}[  q( g \cdot x, \omega)  ] = \rho(g^{-1}) \mathbb{E}_{\omega}[  q( x, \omega)  ]
\end{align}
so that the $G$-equivariance properties of the estimator are recovered after averaging. Note that a weakly equivariant estimator in Eq.~\ref{weak-estimator} will have sample averages that are not guaranteed to be $G$-equivariant. Analogously, we will say that a sample estimator $\hat{f}$ of a $G$-equivariant function $f$ is \textbf{strongly equivariant} if 
\begin{align}\label{strong-estimator}
\forall g \in G, \quad  q( g \cdot x, \omega)   = \rho(g^{-1}) q( x, \omega)  
\end{align}
holds identically. A strongly equivariant estimator in Eq.~\ref{weak-estimator} has sample averages that always satisfies $G$-equivariant condition, regardless of the number of samples. Note that any strongly equivariant estimator is a weakly equivariant estimator, but the converse is not true.

We show in the appendix Sec.~\ref{Toy Models of Equivariant Sampling} that for any compact group $G$ it is possible to construct a strongly equivariant estimator $\hat{f}_{G}$ of $f$ from a weakly equivariant estimator $\hat{f}$ of $f$. Furthermore, the strongly equivariant estimator $\hat{f}_{G}$ is guaranteed to be a better estimator of $f$ than the weakly equivariant estimator $\hat{f}$.

% NOTE: introduce these concepts before constructing the algorithm

\subsection{Constructing Fully Equivariant Version of Action Sampling}
\label{subsec:equiv-action-sampling}

Following the insight, we construct an equivariant action sampling approach based on Implicit Behavior Cloning method (IBC)~\citep{florence_implicit_2021}, an energy-based approach that samples actions from a learned energy model.
We use it to illustrate how to design a sampling-based policy that is not only weakly equivariant but also strongly equivariant in generating action samples.
% The results in Fig.~\ref{fig:results-coordinate-regression}, compare sampling methods \citep{florence_implicit_2021} that are weakly vs strongly equivariant (ours) on a task with $D_{4}$-invariance.

We propose an equivariant action sampling strategy which has two steps can be used to produce action $\va_t$ for one time step. We must enforce equivariance in both the energy function $E(\vs_t,\va_t)$, and (2) equivariance in sampling of actions 
$\va_t \sim \pi ( \va \mid \vs_t)$.
More concretely, we need to make these two steps respect symmetry:
% (1) Constrain the energy function $E(\vs,\va) = E(g \cdot \vs, g \cdot \va)$ to be $G$-invariant. 
% % Nevertheless, an equivariant energy function alone does not embed the equivariance of the policy.
% This only results in \textbf{weak equivariance}: the equivariant energy function only guarantees that the action is equivariant in expectation or under infinite samples, while the set of \textit{finite} samples could break equivariance, especially when the number of samples is small.
% %
% To guarantee \textbf{strong equivariance}, we need (2) additionally ensure the sampling process preserve symmetry, because the CEM sampling samples finite actions from a Gaussian distribution and breaks the symmetries of the problem.
\begin{itemize}[leftmargin=*]
    \item \textit{Step 1: G-Invariant Energy Function: Weak Equivariance.} We enforce the constraint \(E(\vs, \va) = E(g \cdot \vs, g \cdot \va)\) to maintain \(G\)-invariance. Although this condition guarantees equivariance under an infinite sampling regime, finite sample sets can still introduce deviations due to sampling disparities, particularly noticeable when the sample size is small.

    \item \textit{Step 2: Symmetry Preservation in Sampling: Strong Equivariance.} As CEM samples actions from a Gaussian distribution, this randomness can disrupt the underlying symmetries. To counteract this, we need to additionally introduce a symmetry-preserving mechanism in the sampling process.
\end{itemize}

\paragraph{\textbf{G-Augmented Sampling}: Sampling on $G$-orbits.}
We propose a strategy to preserve equivariance in sampling for \textit{strong equivariance}\footnote{We can remove randomness while keeping the correct distribution by reparameterizing the input with standard Gaussian noise.}.
We consider the simplified case with a single time step, so the sampling draws $N$ actions from a random Gaussian distribution $\mathcal{N}(\mu, \sigma^2 I)$, denoted as $\sA = \{ \va_i \}_{i=1}^N$. The score (or ``return'' in RL terminology ) is simply a scalar value $Q(s,a)$, also called a $Q$-value in RL literature \cite{sutton_reinforcement_2018}. To develop analogy with energy models, we will sometime denote the return, as a function of state $s$ and action $a$, as $E(s,a)$. Assuming we only select the best trajectory ($K=1$), we require the sampling algorithm to be equivariant: $a_0 = \argmin_{a} E(s_0, a)$.  % $a_0 = \argmax_{a} Q(s_0, a)$
In other words, if we rotate the state $ s_{0} \to g \cdot s_0$, the selected action is also rotated $a_0 \to g \cdot a_0$.

We can enforce this condition via the following simple strategy: for each single sample $\va$, we augment the action sampling via $g \cdot \va$, i.e., left to the orbit of $G$: $\{  g \cdot \va \mid g \in G \}$.
Thus, the $G$-augmented sample set is $G\sA =\{ g \cdot a_i \mid g \in G \}_{i=1}^N$.
We indicate the sampling strategy as $G$\texttt{-sample}. The equivariance condition for the reward function can be written as
\begin{equation}
    % g \cdot a_0 = g \cdot \argmax_{a\in G\sA} Q(s_0, a) = \argmax_{a\in G\sA} Q(g \cdot s_0, a).
    g \cdot a_0 = g \cdot \argmin_{a\in G\sA} E(s_0, a) = \argmin_{a\in G\sA} E(g \cdot s_0, a).
\end{equation}
For a more detailed account, please refer to Sec~\ref{sec:add_algorithm}.

% \subsection{Evaluation of Equivariant Action Sampling Approach on Coordinate Regression}
\subsection{Evaluation of Equivariant Sampling on Coordinate Regression}

\paragraph{Setup.}
In our experiments, we processed images at a resolution of \(96 \times 96\) pixels. The dataset consisted of 10 training images from either (1) the entire region $[0,96] \times [0, 96]$ or (2) only the first quadrant $[0, 48] \times [0, 48]$. Each image features a single red marker with randomly assigned coordinates (as shown in Fig~\ref{fig:algo-equiv-sampling}), using a fixed random seed to ensure consistency across models.
The coordinates input to the energy function $E(I, v)$ are normalized to $[-1, 1] \times [-1, 1]$.
For the equivariant $E(I, v)$, we use $D_4$-equivariance\footnote{ \tiny
(1) The image $I$ has dimensions $3 \times 96 \times 96$, with group actions involving \textit{i)} spatial rotation of the image, and \textit{ii)} no action on the RGB channels.
(2) The coordinate input $u$ to the energy function is essentially a $2 \times 1 \times 1$ vector (no spatial dimension), with group actions involving \textit{i)} no spatial rotation, and \textit{ii)} standard $G$-representation ($2 \times 2$ rotation matrix).
}.

(1) The \textbf{upper row} of Figure~\ref{fig:results-coordinate-regression} evaluates spatial generalization, presenting the model's test performance on 500 random coordinates. A {\color{blue} blue marker} indicates the model's accurate prediction within a 1-pixel error range.
(2) The \textbf{lower row} shows the learned energy function's landscape across a \(96 \times 96\) grid visualized as a color map, with a test marker fixed at coordinate $(72, 72)$. The color intensity corresponds to the energy levels, guiding the CEM in identifying potential coordinate predictions.

Additionally, we visualized the training data points as \textbf{crosses} ($\times$) within the prediction error (upper row) and energy color map (lower row) and delineated the convex hull of these training points. The convex hull represents the smallest convex set that contains all the training data points and serves as a boundary for evaluating the model's extrapolation capabilities.

\paragraph{Evaluation: Coordinate Regression Spatial Generalization.}
(1) When training across the entire region, the use of equivariance in both energy and sampling resulted in more blue points with errors less than 1 pixel, indicating a well-trained energy function that supports accurate predictions during sampling.
(2) When training was limited to the first quadrant, the non-equivariant model struggled to generalize beyond the convex hull, especially in regions outside \([0, 48] \times [0, 48]\). However, the implementation of a \(G\)-invariant energy function combined with \(G\)-augmented sampling significantly enhanced the model's extrapolation capabilities. This underscores the importance of equivariance in improving model generalization, particularly through the use of augmented sampling.

%%%%%%%%%%%%%%%%
\begin{wrapfigure}{r}{0.45\textwidth}
\vspace{-41pt}
  \begin{center}
    \includegraphics[width=0.43\textwidth]{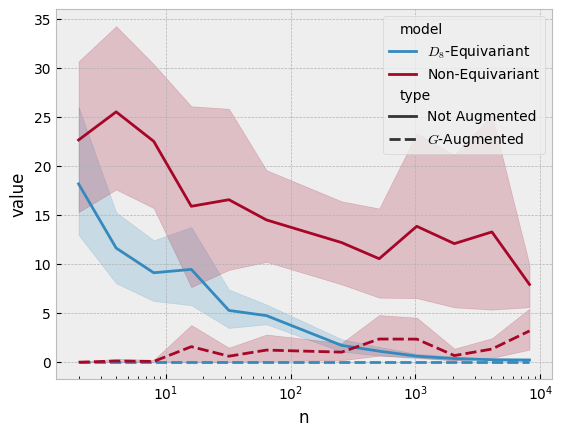}
  \end{center}
  \vspace{-20pt}
  \caption{
\scriptsize
% NOTE: this figure is better to be updated
Measuring the equivariance error of using whether $G$-invariant $E(s,a)$ and whether augment action with $G$.
  }
  \vspace{-20pt}
  \label{fig:demo-ee-sampling}
\end{wrapfigure}
%%%%%%%%%%%%%%%%

\paragraph{Analysis: Equivariance Error in 1-Step CEM.}
To explore the relationship between equivariance and finite samples, we simulate CEM using \textit{untrained} equivariant and non-equivariant energy functions $E(s,a)$, thereby avoiding any learned equivariance from data. We measure the equivariance error under two conditions: (1) using a $D_4$-equivariant ($90^\circ$ rotations and reflections) or non-equivariant energy-based model $E(s,a)$, and (2) employing a $G$-augmented equivariant action sampling strategy. We utilize randomly initialized models and random action samples with $\mu=2$ (shared across four variants but resampled between runs) and average the results over 50 seeds, as shown in Fig~\ref{fig:demo-ee-sampling}. The results indicate that while the equivariant EBM without $G$-augmentation requires more samples to achieve a low equivariance error, the $G$-augmented sampling consistently maintains perfect equivariance. This confirms the superiority of our proposed algorithm over sampling algorithms that lack $G$-action augmentation and those that do not employ a $G$-equivariant model.

%%%%%%%%%%%%%%%%%%%%%%%%%%%%%%%%%%%%%%%%%%%%%%%%%%%%%%%%%%%%%%%%%%%%%%%%%%%%%%%%%%%%%%%%%%%%%%%%%%%%%%
%%%%%%%%%%%%%%%%%%%%%%%%%%%%%%%%%%%%%%%%%%%%%%%%%%%%%%%%%%%%%%%%%%%%%%%%%%%%%%%%%%%%%%%%%%%%%%%%%%%%%%

% \section{Symmetry in Sampling-based Model-based RL Algorithms}
\section{Equivariant Sampling-based Planning Algorithm}
\label{sec:main-algorithm}

%%%%%%%%%%%%%%%%%%%%%%%%%%%%%%%%%%%%
\begin{figure*}[t]
\centering
\includegraphics[width=1.\linewidth]{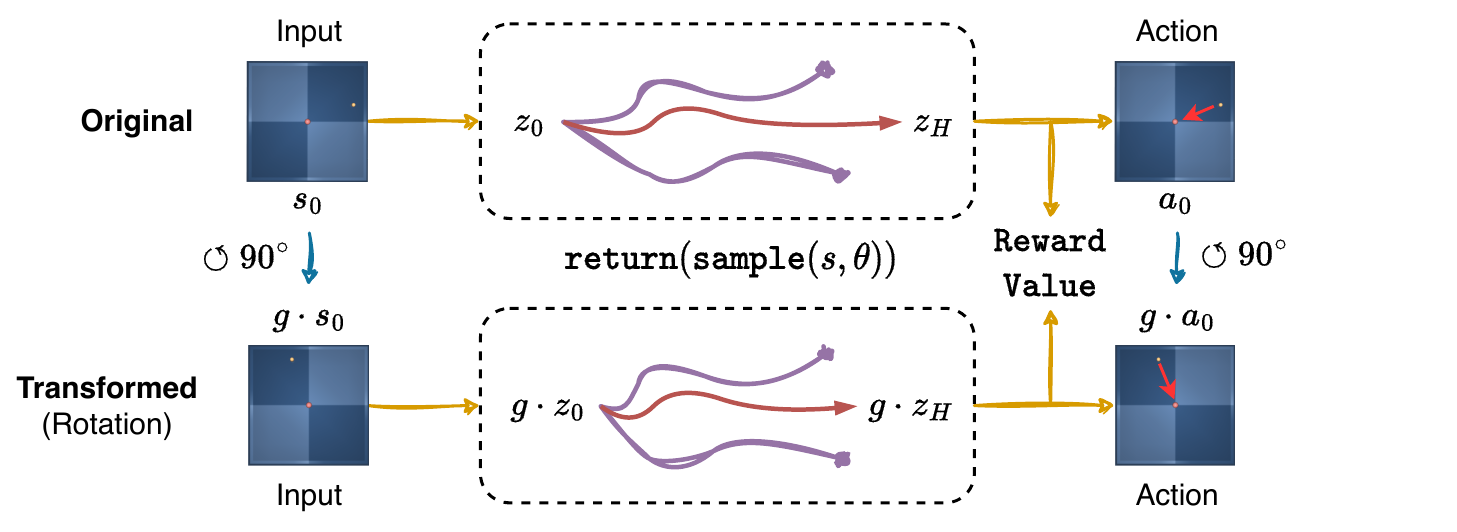}
\vspace{-10pt}
\caption{
\small
The proposed sampling-based planning algorithm $a_0 = \texttt{plan}(s_0)$: if the input state is rotated, the output action should be rotated accordingly. This requires (1) the learned functions to be $G$-equivariant or $G$-invariant networks and (2) a specialized sampling strategy, as introduced in our method.
}
\vspace{-14pt}
\label{fig:algo-equiv-sampling-tdmpc}
\end{figure*}
%%%%%%%%%%%%%%%%%%%%%%%%%%%%%%%%%%%%

In this section, we present an equivariant model-based RL algorithm designed for continuous action spaces, leveraging continuous symmetry through symmetric sampling.  To plan in continuous spaces, we employ \textit{sampling-based} methods such as MPPI \citep{williams_model_2015,williams_information_2017}, extending them to maintain equivariance. Our approach builds on prior work \citep{zhao_integrating_2022} that utilized value-based planning in a discrete state space $\ztwo$ with the discrete group $D_4$, extending these concepts to continuous domains.

The core idea is to ensure that the algorithm $a_t = \texttt{plan}(s_t)$ produces actions that are consistent under transformations, i.e., it is $G$-equivariant: $g \cdot a_t \equiv g \cdot \texttt{plan}(s_t) = \texttt{plan}(g \cdot s_t)$, as illustrated in Figure~\ref{fig:algo-equiv-sampling}. 
This principle is applicable to MDPs with various symmetry groups.

\subsection{Components}

We use TD-MPC \citep{hansen_temporal_2022} as the foundation of our implementation. Here, we introduce the procedure and demonstrating how to incorporate symmetry into sampling-based planning algorithms.

\begin{itemize}[leftmargin=*]
\item \textit{Planning with learned models.} We utilize the MPPI (Model Predictive Path Integral) control method \citep{williams_model_2015,williams_aggressive_2016,williams_model_2017,williams_information_2017}, as adopted in TD-MPC \citep{hansen_temporal_2022}. 
We sample $N$ trajectories with a horizon $H$ using the learned dynamics model, with actions derived from a learned policy, and estimate the expected total return.

\item \textit{Training models.} The learnable components in equivariant TD-MPC include: an \texttt{encoder} that processes input observations, \texttt{dynamics} and \texttt{reward} networks that simulate the MDP, and \texttt{value} and \texttt{policy} networks that guide the planning process.

\item \textit{Loss.} The only requirement is that the loss function is $G$-invariant. The loss terms in TD-MPC include value-prediction MSE loss and dynamics/reward-consistency MSE loss, all of which satisfy invariance.
\end{itemize}

\subsection{Integrating Symmetry}

\citet{zhao_integrating_2022} consider how the Bellman operator transforms under symmetry transformation. For sampling-based methods, one needs to consider how the sampling procedure changes under symmetry transformation. Specifically, under a symmetry transformation, differently sampled trajectories must transform equivariantly. This is shown in Figure~\ref{fig:algo-equiv-sampling}. The equivariance of the transition model in sampling-based approaches to machine learning has also been studied in \citep{park_learning_2022}. There are several components that need $G$-equivariance, and we discuss them step-by-step and illustrate them in Figure~\ref{fig:algo-equiv-sampling}.

\begin{enumerate}[leftmargin=*]
\item \textbf{\texttt{dynamics} and \texttt{reward} model.} In the definition of symmetry in Geometric MDPs (and symmetric MDPs \citep{ravindran2004algebraic,van_der_pol_mdp_2020,zhao_integrating_2022}) in Equation~\ref{eq:symmetry-mdp}, the transition and reward functions are $G$-equivariant and $G$-invariant respectively. Therefore, in implementation, the transition network is deterministic and uses a $G$-equivariant MLP, and the reward network is constrained to be $G$-invariant. Additionally, in implementation, planning is typically performed in latent space, using a latent dynamics model $\bar{f}(\vz, \va) = \vz'$.
\item \textbf{\texttt{value} and \texttt{policy} model.} The optimal value function produces a scalar for each state and is $G$-invariant, while the optimal policy function is $G$-equivariant \citep{ravindran2004algebraic}. If we use $G$-equivariant transition and $G$-invariant reward networks in updating our value function $\mathcal{T}[V_\theta] = \sum_\va R_\theta(\vs,\va) + \gamma \sum_{\vs'} P_\theta(\vs'|\vs, \va) V_\theta(\vs')$, the learned value network $V_\theta$ will also satisfy the symmetry constraint. Similarly, we can extract an optimal policy from the value network, which is also $G$-equivariant \citep{van_der_pol_mdp_2020,wang_mathrmso2-equivariant_2021,zhao_integrating_2022}.
\item \textbf{\texttt{MPC} procedure.} We consider equivariance in the MPC procedure in two parts: \texttt{sample} trajectories from the MDP using learned models, and compute their \texttt{returns}, $\texttt{return}(\texttt{sample}(s, \theta))$. We discuss the invariance and equivariance of it in the next subsection.
\end{enumerate}

We list the equivariance or invariance conditions that each network needs to satisfy. Alternatively, for scalar functions, we can also say they transform under \textit{trivial} representation $\rho_0$ and are thus invariant. All modules are implemented via $G$-steerable equivariant MLPs: $\rho_\text{out}(g) \cdot y = \rho_\text{out}(g) \cdot \texttt{MLP}(x) = \texttt{MLP}(\rho_\text{in}(g) \cdot x)$.

\begin{align}
f_\theta &: \gS \times \gA \to \gS :& \rho_\gS(g) \cdot f_\theta(\vs_{t}, \va_{t}) &=  f_\theta(\rho_\gS(g) \cdot \vs_{t}, \rho_\gA(g) \cdot \va_{t}) \\
R_\theta &: \gS \times \gA \to \sR :& {R}_\theta(\vs_{t}, \va_{t} ) &=  {R}_\theta(\rho_\gS(g) \cdot \vs_{t}, \rho_\gA(g) \cdot \va_{t}) \\
Q_\theta &: \gS \times \gA \to \sR :& {Q}_\theta(\vs_{t}, \va_{t} ) &=  {Q}_\theta(\rho_\gS(g) \cdot \vs_{t}, \rho_\gA(g) \cdot \va_{t}) \\
\pi_\theta &: \gS \to \gA  :& \rho_\gA(g) \cdot {\pi}_\theta(\cdot \mid \vs_{t}) &=  {\pi}_\theta(\cdot \mid \rho_\gS(g) \cdot \vs_{t})
\end{align}

\subsection{Equivariance of \texttt{MPC}}

Analogous to equivariant action selection for single-step case, we constrain the underlying MPC planner to be equivariant. We use MPPI (Model Predictive Path Integral) \citep{williams_model_2015,williams_model_2017}, which has been used in TD-MPC for action selection. An MPPI procedure samples multiple $H$-horizon trajectories $\{ \tau_i \}$ from the current state $\vs_t$ using the learned models. We use $\texttt{sample}$ to refer to the procedure: $\tau_i \equiv \texttt{sample}(\vs_t ; f_\theta, R_\theta, Q_\theta, \pi_\theta) = (\vs_t, \va_t, \vs_{t+1}, \va_{t+1}, \ldots, \vs_{t+H})$. Another procedure $\texttt{return}$ computes the accumulated return, evaluating the value of a trajectory for top-$k$ trajectories:

\begin{equation}
    \texttt{return}({\tau}) = \mathbb{E}_{\tau}\left[\gamma^H Q_\theta\left(\vs_H, \va_H\right)+\sum_{t=0}^{H-1} \gamma^t R_\theta\left(\vs_t, \va_t\right)\right] = \mathbb{E}_{\tau}\left[ U(\vs_{1:H}, \va_{1:H-1}) \right]
\end{equation}

A trajectory is transformed element-wise by $g$: $g\cdot \tau_i = (g \cdot \vs_t, g \cdot \va_t, g \cdot \vs_{t+1}, g \cdot \va_{t+1}, \ldots, g \cdot \vs_{t+H})$. However, since $\mu$ and $\sigma$ in action sampling are \textit{not state-dependent}, the MPPI \texttt{sample} does not exactly preserve equivariance: rotating the input does not \textit{deterministically} guarantee a rotated output, similar to CEM. Thus, we can (1) constrain return to be $G$-invariant and (2) use $G$ to augment the sampling of action sequence $(\va_t, \ldots, \va_{t+H})$.

\begin{proposition}
The \texttt{return} procedure is $G$-invariant, and the $G$-augmented $G$\texttt{-sample} procedure that augment $\sA$ using transformation in $G$ is $G$-equivariant when $K=1$.
\end{proposition}

We further explain in Appendix~\ref{sec:add_algorithm}. In summary, for sampling and computing return, the sampling procedure satisfies the following conditions, indicating that the procedure $\texttt{return}(G\texttt{-sample}(s, \theta))$ is invariant, i.e., not changed under group transformation for any $g$. We use $\texttt{return}(\tau_i)$ to indicate the return of a specific trajectory $\tau_i$ and $g\cdot \tau_i$ to denote group action on it.

\begin{align}
G\texttt{-sample} &: \vs_t, \theta \mapsto \tau_i  :& g \cdot \tau_i &\sim G\texttt{-sample}(g \cdot \vs_t ; f_\theta, R_\theta, Q_\theta, \pi_\theta) \\
\texttt{return} &: \tau_i \mapsto \sR  :& \texttt{return}(\tau_i) &= \texttt{return}(g \cdot \tau_i)
\end{align}

%%%%%%%%%%%%%%%%%%%%%%%%%%%%%%%%%%%%
\begin{figure*}[t]
\centering
%\vspace{-20pt}
\subfigure{
\includegraphics[width=.9\linewidth]{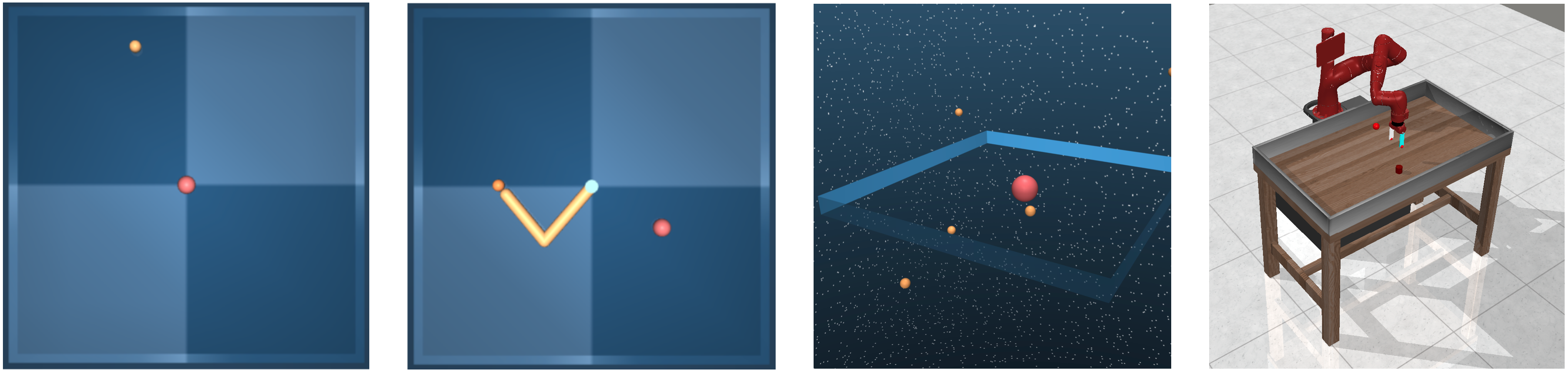}
}
% \vspace{-15pt}
\centering
\caption{
\small
Tasks used in experiments:
% (1) \texttt{PointMass} in 2D, (2) \texttt{Reacher}, (3) \texttt{PointMass} customized 3D version, and (4) \texttt{PointMass} customized 3D version with multiple particles to control.
(1) \texttt{PointMass} in 2D, (2) \texttt{Reacher}, (3) Customized 3D version of \texttt{PointMass} with multiple particles to control, and (4) MetaWorld task to reach an object with gripper.
}
 \vspace{-10pt}
\label{fig:env-collection}
\end{figure*}
%%%%%%%%%%%%%%%%%%%%%%%%%%%%%%%%%%%%

%%%%%%%%%%%%%%%%%%%%%%%%%%%%%%%%%%%%%%%%%%%%%%%%%%%%%%%%%%%%%%%%%%%%%%%%%%%%%%%%%%%%%%%%%%%%%%%%%%%%%%
%%%%%%%%%%%%%%%%%%%%%%%%%%%%%%%%%%%%%%%%%%%%%%%%%%%%%%%%%%%%%%%%%%%%%%%%%%%%%%%%%%%%%%%%%%%%%%%%%%%%%%

% \section{Experiments}
\section{Evaluation: Sampling-Based Planning}

In this section, we present the setup and results for our proposed sampling-based planning algorithm: the equivariant version of TD-MPC.
Additional details and results are available in \edit{Appendix~\ref{sec:add_results}}.

\subsection{Experimental Setup}

\paragraph{Tasks.}
We verify the algorithm on a few selected and customized tasks using the DeepMind Control suite (DMC)~\citep{tassa_deepmind_2018}, visualized in Figure~\ref{fig:env-collection}.
One task is a 2D particle moving in $\sR^2$, \texttt{PointMass}.
We customize tasks based on it: (1) 3D particle moving in $\sR^3$ (disabled gravity), and (2) 3D $N$-point moving that has several particles to control simultaneously.
The goal is to move particle(s) to a target position.
We also experiment with tasks on a two-link arm, \texttt{Reacher} (easy and hard), where the goal is to move the end-effector to a random position in a plane. \texttt{Reacher} \texttt{Easy} and \texttt{Hard} are top-down tasks where the goal is to reach a random 2D position.
If we rotate the MDP, the angle between the first and second links is not affected, i.e., it is $G$-invariant.
The first joint and the target position are transformed under rotation, so we set it to $\rho_1$ standard representation (2D rotation matrices).
The system has $\otwo$ rotation and reflection symmetry, hence we use $D_8$ and $D_4$ groups. We also use MetaWorld tabletop manipulation \citep{yu_meta-world_2019}.
The action space is 3D gripper movement $(\Delta x, \Delta y, \Delta z)$ and 1D openness.
The state space includes (1) gripper position, (2) 3D position plus 4D quaternion of at most 2 relevant objects, and (3) 3D randomized goal position, depending on tasks.
If we consider tasks with gravity, the MDP itself should exhibit $\sotwo$ symmetry about the gravity axis.
In implementation, the symmetry also depends on the data distribution, so we make the origin at the workspace center and the gripper initialized at the origin, so the task respects rotation equivariance around the origin.
We add $\sotwo$ equivariance to the algorithm about the gravity axis.

%%%%%%%%%%%%%%%%%%%%%%%%%%%%%%%%%%%%
\begin{figure*}[t]
\centering
\subfigure{
\includegraphics[width=1.\linewidth]{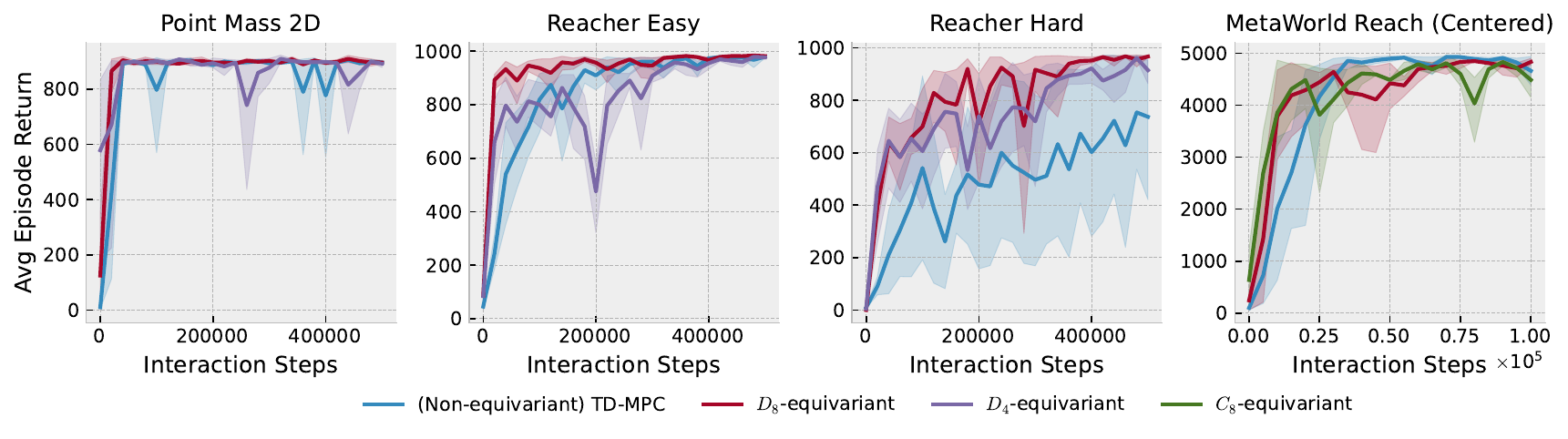}
}
\vspace{-5pt}
\subfigure{
\includegraphics[width=1.\linewidth]{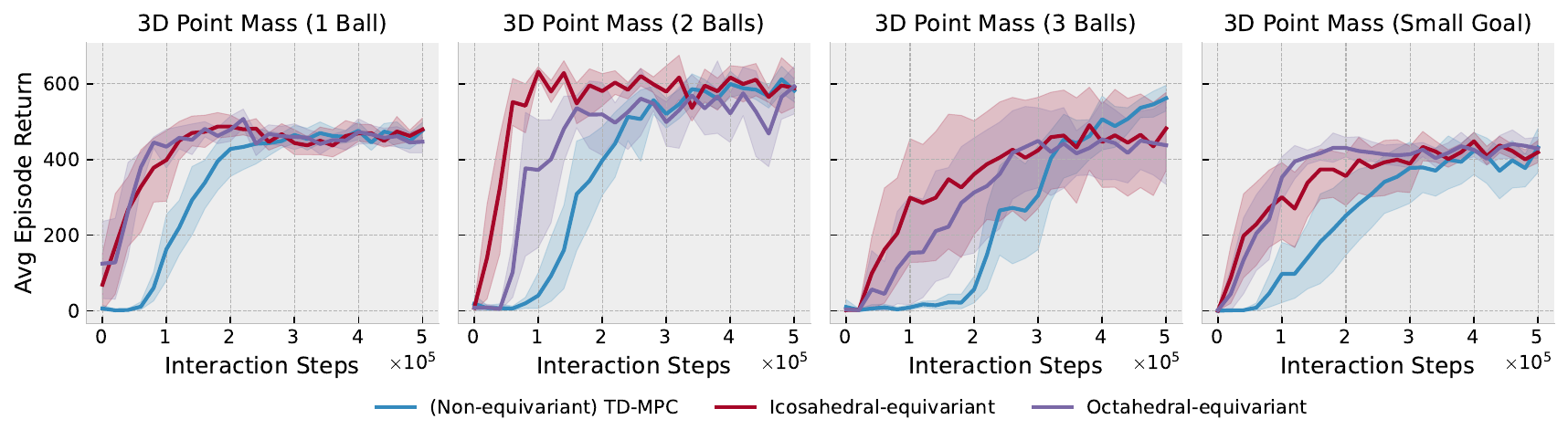}
}
 \vspace{-18pt}
\centering
\caption{
\small
(Upper) Results on \texttt{PointMass}, \texttt{Reacher}, and MetaWorld \texttt{Reach} task.
(Lower) A set of customized 3D $N$-ball \texttt{PointMass} tasks, with $N=1,2,3$, and a customized 3D \texttt{PointMass} with a smaller target.
}
\vspace{-14pt}
\label{fig:env-reward-curves-two-row}
\end{figure*}
%%%%%%%%%%%%%%%%%%%%%%%%%%%%%%%%%%%%

\paragraph{Experimental setup.}
We compare against the non-equivariant version of TD-MPC \citep{hansen_temporal_2022}.
By default, we make all components equivariant as described in the algorithm section.
In \edit{Sec~\ref{subsec:add-results-appendix}}, we include ablation studies for disabling or enabling each equivariant component.

The training procedure follows TD-MPC \citep{hansen_temporal_2022}.
We use the state as input and for equivariant TD-MPC, we divide the original hidden dimension by $\sqrt{N}$, where $N$ is the group order, to keep the number of parameters roughly equal between the equivariant and non-equivariant versions.
We mostly follow the original hyperparameters except for \texttt{seed\_steps}.
We use $5$ random seeds for each method.

\paragraph{Algorithm setup: equivariance.}
We use discretized subgroups in implementing $G$-equivariant MLPs with the \texttt{escnn} package \citep{weiler_general_2021}, which are more stable and easier to implement than continuous equivariance.
For the 2D case, we use $\otwo$ subgroups: dihedral groups $D_4$ and $D_8$ ($4$ or $8$ rotation components), or rotation group $C_8$ ($45^\circ$ rotations).
For the 3D case, we use the icosahedral and octahedral groups, which are finite subgroups of $\sothree$ with orders $60$ and $24$ respectively. On \texttt{Reacher} tasks, we also compare against a \textit{planning-free} baseline by removing MPPI planning with the learned model and only keeping the policy learning, as shown in Fig~\ref{fig:ablation-planning-component}.

\todozlf{add some explanation from rebuttal}

\subsection{Results}

Figures~\ref{fig:env-reward-curves-two-row} (upper and lower rows) present the reward curves, demonstrating that our equivariant methods can achieve near-optimal performance 2 to 3 times faster in terms of training interaction steps for several tasks.
For the default 2D \texttt{PointMass} task, the $D_8$-equivariant version learns slightly faster than the non-equivariant version.
In the \texttt{Reacher} task, as shown in the lower part of Figure~\ref{fig:env-reward-curves-two-row}, the $D_8$-equivariant version significantly outperforms the non-equivariant TD-MPC, especially in the \texttt{Hard} domain.
The $D_4$-equivariant version also performs better than the baseline, though not as well as $D_8$.
The rightmost plot illustrates the MetaWorld \texttt{Reach} task, which involves reaching a button on a desk using a parallel gripper.
We added $\sotwo$ equivariance to the algorithm about the gravity axis and evaluated the $C_8$ and $D_8$-equivariant versions, both of which demonstrated more efficient learning.

In the \texttt{Reacher} tasks, we also compared against a \textit{planning-free} baseline by removing MPPI planning with the learned model and retaining only policy learning, as shown in Figure~\ref{fig:ablation-planning-component}.
This approach is effectively similar to the DDPG algorithm \citep{lillicrap_ddpg_2016}.

We designed a set of more challenging 3D versions of \texttt{PointMass} and used $\sothree$ subgroups to implement 3D equivariant versions of TD-MPC, utilizing icosahedral- and octahedral-equivariant MLPs.
Figure~\ref{fig:env-reward-curves-two-row} (lower) shows tasks with $N=1,2,3$ balls in 3D \texttt{PointMass}, and the rightmost figure depicts a 1-ball 3D version with a smaller target ($0.02$ compared to $0.03$ in the $N$-ball version).

We found that the icosahedral (order $60$) equivariant TD-MPC consistently learns faster and uses fewer samples to achieve the best rewards compared to the non-equivariant version.
The octahedral (order $24$) equivariant version also performs similarly.
Interestingly, the best absolute rewards in the 1-ball case are lower than in the 2- and 3-ball cases, which may be due to the higher possible return from having 2 or 3 balls that can reach the goal.

With higher-order 2D discrete subgroups, the performance plateaus but computational costs increase, so we use up to $D_8$.
We also find TD-MPC is especially sensitive to a hyper-parameter \texttt{seed\_steps} that controls the number of warm-up trajectories.
In contrast, our equivariant version is robust to it and sometimes learns better with less warm-up.
In the shown curves, we do not use warm-up across non-equivariant and equivariant ones and present additional results in \edit{Appendix~\ref{sec:add_results}}.
Most of these tasks are goal-reaching and have optimal rewards, thus the number of transitions used to reach near-optimal is a proxy of sample efficiency.
The reward curves show the superiority of our equivariant sampling-based approach.

%%%%%%%%%%%%%%%%%%%%%%%%%%%%%%%%%%%%%%%%%%%%%%%%%%%%%%%%%%%%%%%%%%%%%%%%%%%%%%%%%%%%%%%%%%%%%%%%%%%%%%
%%%%%%%%%%%%%%%%%%%%%%%%%%%%%%%%%%%%%%%%%%%%%%%%%%%%%%%%%%%%%%%%%%%%%%%%%%%%%%%%%%%%%%%%%%%%%%%%%%%%%%

\section{Conclusion and Discussion}

This work introduces a two-step approach to preserve symmetry in sampling-based planning and control for continuous tasks. Using equivariant sampling, our method improves decision-making efficiency and performance in various control environments. Our findings highlight the benefits of integrating symmetry into sampling-based model-based RL algorithms, enhancing current practices and opening avenues for future research in continuous control and robotics applications.

\noindent
{\small \textbf{Acknowledgements}
This work was partially supported by NSF Grant 2107256. Owen Howell is indebted to the National Science Foundation Graduate Research Fellowship Program (NSF-GRFP) for financial support. Robin Walters is supported by NSF DMS-2134178.}

%%%%%%%%%%%%%%%%%%%%%%%%%%%%%%%%%%%%%%%%%%%%%%%%%%%%%%%%%%%%%%%%
%% Bibliography
%%%%%%%%%%%%%%%%%%%%%%%%%%%%%%%%%%%%%%%%%%%%%%%%%%%%%%%%%%%%%%%%
\bibliography{zotero-linfeng,main,refs}
\bibliographystyle{unsrtnat}
% \bibliographystyle{abbrvnat}

%%%%%%%%%%%%%%%%%%%%%%%%%%%%%%%%%%%%%%%%%%%%%%%%%%%%%%%%%%%%%%%%
%% Appendices
%%%%%%%%%%%%%%%%%%%%%%%%%%%%%%%%%%%%%%%%%%%%%%%%%%%%%%%%%%%%%%%%
\newpage

\appendix

% > add table of contents
\tableofcontents

% > resume
\addtocontents{toc}{\protect\setcounter{tocdepth}{2}}

\listoftodos

%%%%%%%%%%%%%%%%%%%%%%%%%%%%%%%%%%%%%%%%%%%%%%%%%%%%%%%%%%%%%%%%%%%%%%%%%%%%%%%%%%%%%%%%%%%%%%%%
%%%%%%%%%%%%%%%%%%%%%%%%%%%%%%%%%%%%%%%%%%%%%%%%%%%%%%%%%%%%%%%%%%%%%%%%%%%%%%%%%%%%%%%%%%%%%%%%

\section{Outline}

\todoil{update appendix outline}

The appendix is organized as follows: (1) additional discussion, including related work and theoretical background, (2) theory, derivation, and proofs, (3) implementation details and further empirical results, and (4) additional mathematical background.

%%%%%%%%%%%%%%%%%%%%%%%%%%%%%%%%%%%%%%%%%%%%%%%%%%%%%%%%%%%%%%%%%%%%%%%%%%%%%%%%%%%%%%%%%%%%%%%%
%%%%%%%%%%%%%%%%%%%%%%%%%%%%%%%%%%%%%%%%%%%%%%%%%%%%%%%%%%%%%%%%%%%%%%%%%%%%%%%%%%%%%%%%%%%%%%%%

\section{Additional Discussion}
\label{sec:add_discussion}

% \begin{itemize}
% \color{brown}
%     \item discuss - what symmetry
%     \item additional related work
%     \item discussion of limitations / future work / other directions
% \end{itemize}

%%%%%%%%%%%%%%%%%%%%%%%%%%%%%%%%%%%%%%%%%%%%%%%%%%%%%%%%%%%%%%%%%%%%%%%%%%%%%%%%%%%%%%%%%%%%%%%%
% \subsection{Discussion: Symmetry in MDPs}
\subsection{Discussion: Symmetry in Decision-making}

%\begin{itemize}
%\color{brown}
%    \item what would be the symmetry here?
%\end{itemize}

In this work, we study the Euclidean symmetry $\mathrm{E}(d)$ from geometric transformations between \textit{reference frames}.
This is a specific set of symmetries that an MDP can have -- isometric transformations of Euclidean space $\sR^d$, such as the distance is preserved.
This can be viewed as a special case under the framework of MDP homomorphism, where symmetries relate two different MDPs via MDP \textit{homomorphism} (or more strictly, \textit{isomorphism}).
We refer the readers to \citep{ravindran2004algebraic} for more details.
We also discuss symmetry in other related fields.

Classic planning algorithms and model checking have leveraged the use of symmetry properties, \citep{fox_detection_1999, fox_extending_2002, pochter_exploiting_2011, domshlak_enhanced_nodate, shleyfman_heuristics_2015, holldobler_empirical_2015, sievers_structural_nodate, sievers_theoretical_2019, fiser_operator_2019} as evident from previous research. In particular, \citet{zinkevich_symmetry_2001} demonstrate that the value function of an MDP is invariant when symmetry is present. However, the utilization of symmetries in these algorithms presents a fundamental problem since they involve constructing equivalence classes for symmetric states, which is difficult to maintain and incompatible with differentiable pipelines for representation learning. \citet{narayanamurthy_hardness_2008} prove that maintaining symmetries in trajectory rollout and forward search is intractable (NP-hard). To address the issue, recent research has focused on state abstraction methods such as the coarsest state abstraction that aggregates symmetric states into equivalence classes studied in MDP homomorphisms and bisimulation \citep{ravindran2004algebraic, ferns_metrics_2004, li_towards_2006}. However, the challenge lies in that these methods typically require perfect MDP knowledge and do not scale well due to the complexity of constructing and maintaining abstraction mappings \citep{van2020mdp}. To deal with the difficulties of symmetry in forward search, recent studies have integrated symmetry into reinforcement learning based on MDP homomorphisms \citep{ravindran2004algebraic}, including \citet{van2020mdp} that integrate symmetry through an equivariant policy network. Furthermore, \citet{mondal_group_2020} previously applied a similar idea without using MDP homomorphisms. \citet{park_learning_2022} learn equivariant transition models, but do not consider planning, and \citet{zhao_toward_2022} focuses on permutation symmetry in object-oriented transition models. Recent research by \citep{zhao_integrating_2022,zhao_scaling_2023} on 2D discrete symmetry on 2D grids has used a value-based planning approach.

% > brief version
%\paragraph{Benefits of symmetry in continuous control.}
%\textbf{Benefits of symmetry in continuous control.}
% \textbf{Remark.}
There are some benefits of explicitly considering symmetry in continuous control.
The possibility of hitting orbits is negligible, so there is no need for orbit-search on symmetric states in forward search in continuous control.
Additionally, the planning algorithm implicitly plans in a smaller continuous MDP $\mathcal{M} / G$ \citep{ravindran2004algebraic}.
Furthermore, from equivariant network literature \citep{elesedy_provably_2021}, the generalization gap for learned equivariant policy and value networks are smaller, which allows them to generalize better.

%%%%%%%%%%%%%%%%%%%%%%%%%%%%%%%%%%%%%%%%%%%%%%%%%%%%%%%%%%%%%%%%%%%%%%%%%%%%%%%%%%%%%%%%%%%%%%%%
% \subsection{Additional Related Work}

% NOTE: the original additional related work has been merged into the main paper; the original version sees /previous/appendix_rlc2024.tex

%%%%%%%%%%%%%%%%%%%%%%%%%%%%%%%%%%%%%%%%%%%%%%%%%%%%%%%%%%%%%%%%%%%%%%%%%%%%%%%%%%%%%%%%%%%%%%%%
\subsection{Limitations and Future Work}

% NOTE: moved previous discussion on GMDP examples to /unused/unused.tex

Although Euclidean symmetry group is infinite and seems huge, it does not guarantee significant performance gain in all cases.
Our theory helps us understand when such Euclidean symmetry may not be very beneficial
The key issue is that when a robot has kinematic constraints, Euclidean symmetry does not change those features, which means that equivariant constraints cannot share parameters and reduce dimensions.
We empirically show this on using local vs. global reference frame in the additional experiment in Sec~\ref{sec:add_results}.
For further work, one possibility is to explicit consider constraints while keep using global positions.

%%%%%%%%%%%%%%%%%%%%%%%%%%%%%%%%%%%%%%%%%%%%%%%%%%%%%%%%%%%%%%%%%%%%%%%%%%%%%%%%%%%%%%%%%%%%%%%%
\section{Mathematical Background}
\subsection{Background for Representation Theory and $G$-steerable Kernels}

We establish some notation and review some elements of group theory and representation theory. For a comprehensive review of group theory and representation theory, please see \citep{Serre_2005}. The identity element of any group $G$ will be denoted as $e$. We will always work over the field $\mathbb{R}$ unless otherwise specified.

\subsection{Group Definition}
A group is a non-empty set equipped with an associative binary operation $\cdot: G \times G \rightarrow G$ where $\cdot$ satisfies
\begin{align*}
 & \text{Existence of identity: } \exists e \in G, \text{ s.t. } \forall g\in G, \enspace e \cdot g = g \cdot e = g \\
& \text{Existence of inverse: } \forall g\in G, \exists g^{-1} \in G \text{ s.t. } g \cdot g^{-1} = g^{-1} \cdot g = e
\end{align*}

For a complete reference on group theory, please see \cite{Zee_2016}.

% \subsection{Group Representations}
\subsubsection{Group Representations}
A group is an abstract object. Oftentimes, when working with groups, we are most interested in group \emph{representations}.
Let $V$ be a vector space over $\mathbb{C}$. A \emph{representation} $(\rho , V)$ of $G$ is a map $\rho : G \rightarrow \Hom[V,V]$ such that 
\begin{align*}
\forall g , g' \in G, \enspace \forall v\in V, \quad   \rho( g \cdot g' )v =  \rho( g ) \cdot \rho(  g' )v
\end{align*}
Concisely, a group representation is a embedding of a group into a set of matrices. The matrix embedding must obey the multiplication rule of the group. Over $\mathbb{R}$ and $\mathbb{C}$ all representations break down into irreducible representations \cite{Serre_2005}. We will denote the set of irreducible representations of a group $G$ and $\hat{G}$.

% \subsection{Group Actions}\label{Section:Group Actions}
\subsubsection{Group Actions}\label{Section:Group Actions}

Let $\Omega$ be a set. A group action $\Phi$ of $G$ on $\Omega$ is a map $\Phi : G \times \Omega \rightarrow \Omega$ which satisfies 
\begin{align*}
    &\text{Identity: } \forall \omega \in \Omega, \quad \Phi(e , \omega )  =  \omega \\
    &\nonumber \text{Compositionality: }\forall g_{1},g_{2} \in G,\enspace \forall \omega \in \Omega, \quad  \Phi( g_{1}g_{2} , \omega  ) = \Phi(g_{1} , \Phi(g_{2}, \omega ))
\end{align*}
We will often suppress the $\Phi$ function and write $\Phi( g , \omega ) = g \cdot \omega$.

\begin{center}\label{Diagram:G-Equivarient_Map}
    \begin{tikzcd}\centering
        &\Omega \arrow{d}{\Phi(g,\cdot)} \arrow{r}{ \Psi } & \Omega' \arrow{d}{ \Phi'(g, \cdot ) }  \\
        & \Omega \arrow{r}{\Psi }  & \Omega'
    \end{tikzcd}
    \captionof{figure}{Commutative Diagram For $G$-equivariant function: Let $\Phi(g, \cdot ): G \times \Omega \rightarrow \Omega$ denote the action of $G$ on $\Omega$. Let $\Phi'(g, \cdot ): G \times \Omega' \rightarrow \Omega'$ denote the action of $G$ on $\Omega'$ The map $\Psi: \Omega \rightarrow \Omega'$ is $G$-equivariant if and only if the following diagram is commutative for all $g\in G$.    }
\end{center}

Let $G$ have group action $\Phi$ on $\Omega$ and group action $\Phi'$ on $\Omega'$. A mapping $\Psi : \Omega \rightarrow \Omega'$ is said to be $G$-equivariant if and only if
\begin{align}\label{Equation:G_Equivariece_Def}
    \forall g \in G, \forall \omega \in \Omega, \quad	 \Psi(  \Phi( g , \omega )  ) = \Phi'( g , \Psi(\omega) )
\end{align}
Diagrammatically, $\Psi$ is $G$-equivariant if and only if the diagram \ref{Diagram:G-Equivarient_Map} is commutative.

%%%%%%%%%%%%%%%%%%%%%%%%%%%%%%%%%%%%%%%%%%%%%%%%%%%%%%%%%%%%%%%%%%%%%%%%%%%%%%%%%%%%%%%%%%%%%%%%
%%%%%%%%%%%%%%%%%%%%%%%%%%%%%%%%%%%%%%%%%%%%%%%%%%%%%%%%%%%%%%%%%%%%%%%%%%%%%%%%%%%%%%%%%%%%%%%%

\section{Theory and Proofs}
% \section{Proofs for Equivariance}
\label{sec:theory-appendix}

% NOTE: major change logs
% completely removed LQR / linearized dynamics related stuff
% moved Owen's sections on equivariant sampling here

This section includes more insights and explanation of equivariant sampling and its benefits, as well as the equivariance properties of Geometric MDP.

\subsection{Toy Models of Equivariant Sampling}
\label{Toy Models of Equivariant Sampling}

Let us consider a toy model of equivariant sampling. This will illustrate the importance of symmetry considerations in sampling methods. Specifically, this example illustrates that if sampling is performed incorrectly, the finite sample averages will not have the same symmetries as the infinite sample average. We show how the desired symmetries can be recovered via a `group averaging'. Let $f : \mathbb{R} \rightarrow \mathbb{R}$ be any smooth function. Let us define the `energy' function $H: \mathbb{R}^{d} \rightarrow \mathbb{R}$ as
\begin{align*}
H(x)  = \mathbb{E}_{\omega}[  f( \omega^{T}x ) ]
\end{align*}
where the random vector $\omega \sim \mathcal{N}(0 , \mathbb{I}_{d} )$ is drawn from a normal distribution with zero mean and identity matrix covariance $\mathbb{I}_{d}$. The random variable $\omega$ is isotropic and both $\omega$ and $O\omega$ are drawn from the same distribution for any orthogonal matrix $O$. Thus, we have that
\begin{align*}
\forall O \in O(d), \quad H(O\cdot x) = \mathbb{E}_{\omega}[  f( \omega^{T}Ox ) ] = \mathbb{E}_{\omega}[  f( (O\omega)^{T}x ) ] = \mathbb{E}_{\omega}[  f( \omega x ) ] = H(x)
\end{align*}
where we have used the fact that the random variable $\omega$ satisfies the property $\omega = O \omega$. Thus, $H(Ox) = H(x)$ is a left $O(d)$-invariant quantity. Now, suppose that we try to approximate $H(x)$ by random sampling. In the naive approach to estimation of $H(x)$, we can drawn $m$ iid sample points $\omega_{1},\omega_{2}, ... , \omega_{m}$ iid from $\mathcal{N}(0,\mathbb{I}_{d})$ and estimate the function $H(x)$ as
\begin{align*}
\hat{H}(x) = \frac{1}{M} \sum_{m=1}^{M} f( \omega_{i} x )
\end{align*}
However, this approximation $\hat{H}(x)$ does not need to satisfy the original symmetry property and $\hat{H}(Ox) = \hat{H}(x)$ is not guaranteed to hold. Because we know that the true function $H(x)$ is left $O(d)$-invariant, it seems like we should be able to construct and estimate to $H(x)$ that is always left $O(d)$-invariant. Let $G$ be a compact group, we can always `symmetrize' the sample estimate $\hat{H}(x)$ via
\begin{align*}
\hat{H}_{G}(x) = \int_{g \in O(d) }dg \text{ }  \hat{H}( g \cdot x) 
\end{align*}
The symmetrized $\hat{H}_{G}(x)$ is then guaranteed to satisfy $\hat{H}_{G}(g \cdot x) = H_{G}(x)$ for all $g\in O(d)$. Furthermore, the symmetrized estimate $\hat{H}_{G}$ is always guaranteed to be a better estimate of $H$ than $\hat{H}$. To see this, note that, for any function $F: G \rightarrow \mathbb{C}$,
\begin{align*}
| \int_{ g\in G }dg \text{ } F(g)  | \leq  \int_{ g\in G }dg \text{ } | F(g)  |
\end{align*}
holds via the triangle inequality. Thus, letting $F(g) = H(x) - \hat{H}( g \cdot x)$ we have that
\begin{align*}
| \int_{ g\in G }dg \text{ } H( x) - \hat{H}( g \cdot x)  | \leq  \int_{ g\in G }dg \text{ } | H( x) - \hat{H}( g \cdot x) | 
\end{align*}
Ergo, by definition of the symmetrized estimate $\hat{H}^{G}$, we have that
\begin{align*}
\int_{ g\in G }dg \text{ } | H(g \cdot x) - \hat{H}^{G}(g \cdot x)  | \leq  \int_{ g\in G }dg \text{ } | H( g \cdot x) - \hat{H}( g \cdot x) | 
\end{align*}
so that the error $| H(x) - \hat{H}^{G}(x)  |$ is always less than the error $| H(x) - \hat{H}(  x) | $ when averaged on $G$ orbits. This is example is of course artificial. However, the sampling methodology developed in (ref main text) ensures that sample averages always have the same symmetries of the true energy functional is based on the same idea. It is easy to see that this can be extended from $G$-invariant to $G$-equivariant functions by modifying the averaging operator,
\begin{align*}
\int_{g \in O(d) }dg \text{ }  \hat{H}( g \cdot x)   \rightarrow \int_{g \in O(d) }dg \text{ }  \rho(g^{-1})\hat{H}( g \cdot x) 
\end{align*}

\subsection{Equivariant Sampling Is Always Better}

Let $G$ be a compact group. Suppose that we have an MDP with symmetry with optimal policy $\pi^{\star}(a|s)$. Using a result of \citep{van_der_pol_mdp_2020}, the optimal policy satisfies the relation $ \forall g\in G, \quad \pi^{\star}(ga \mid gs) =\pi^{\star}(a \mid s)  $.  

Let us suppose that we have an learning policy $\pi$, which may or may not satisfy the $ \pi(ga|gs) = \pi(a|s)$ condition derived in \citep{van_der_pol_mdp_2020}. Given any policy $\pi(a|s)$ we can always `symmetrize' the policy by defining a new policy $\Pi_{G}[\pi] : S \rightarrow A$ defined as
\begin{align*}
\Pi_{G}[\pi](a|s) = \int_{g \in G}dg \text{ } \pi(g\cdot a | g \cdot s) 
\end{align*}
Then, the symmetrized policy $\Pi_{G}[\pi]$ satisfies the condition
\begin{align*}
\forall g \in G, \quad  \Pi_{G}[\pi]( g \cdot a| g \cdot s) = \Pi_{G}[\pi](a|s)  
\end{align*}
The operator $\Pi_{G}$ can be viewed as a operator which takes as input an arbitrary policy $\pi$ and returns a policy $\Pi_{G}[\pi]$ which is $G$-invariant.
Under the assumption of un-biasedness, the symmetrized policy $\Pi_{G}[\pi]$ is always better than the policy $\pi$. To see this, let $D$ be a metric on the action space and consider the $G$-averaged error 
\begin{align*}
\int_{g \in G} dg \text{ } D( \pi^{\star}(ga|gs) , \pi(ga|gs) ) 
\end{align*}
where $\pi^{\star}(ga|gs) = \pi^{\star}(a|s)$ is the true optimal policy. Now, using the triangle inequality, we have that
\begin{align*}
D( \pi^{\star}(a|s) , \int_{g \in G} dg \text{ }  \pi(ga|gs) ) \leq \int_{g \in G} dg \text{ } D( \pi^{\star}(ga|gs) , \pi(ga|gs) ) 
\end{align*}
Thus, using the definition of the $G$-averaged policy, we have that
\begin{align*}
D( \pi^{\star}(a|s) , \Pi_{G}[\pi](a|s)  ) \leq \int_{g \in G} dg \text{ } D( \pi^{\star}(ga|gs) , \pi(ga|gs) ) 
\end{align*}
Using the fact that both $\pi^{\star} $ and $ \Pi_{G}[\pi]$ are $G$-invariant, we can rewrite this as,
\begin{align*}
\int_{g \in G} dg \text{ } D( \pi^{\star}(ga|gs) , \Pi_{G}[\pi](ga|gs)  ) \leq \int_{g \in G} dg \text{ } D( \pi^{\star}(ga|gs) , \pi(ga|gs) ) 
\end{align*}
Ergo, the $G$-averaged policy $\Pi_{G}[\pi]$ is always closer to the true policy than the policy $\pi$. Thus, a arbitrary policy is always worse than its symmetrized counterpart.

%%%%%%%%%%%%%%%%%%%%%%%%%%%%%%%%%%%%%%%%%%%%%%%%%%%%%%%%%%%%%%%%%%%%%%%%%%%%%%%%%%%%%%%%%%%%%%%%
% \subsection{Proofs: Theorem 1 and 2}
\subsection{Theorem 1: Equivariance in Geometric MDPs}

\todoil{check the naming, e.g., GMDP}

\textbf{\textit{Theorem 1}}
\textit{The Bellman operator of a GMDP is equivariant under Euclidean group $\mathrm{E}(d)$.}

\textit{Proof.}
The Bellman (optimality) operator is defined as
\begin{equation}
    \gT [V](\vs)  := \max_{\va} R(\vs, \va) + \int d\vs' P(\vs' \mid \vs, \va) V(\vs'),
\end{equation}
where the input and output of the Bellman operator are both value function $V: \gS \to \sR$.
The theorem directly generalizes to $Q$-value function.

Under group transformation $g$, a feature map (field) $f: X \to \sR^{c_\text{out}}$ is transformed as:
\begin{equation}
	\left[L_g f\right](x)=\left[f \circ g^{-1}\right](x)= \rho_\text{out}(g) \cdot f\left(g^{-1} x\right),
\end{equation}
where $\rho_\text{out}$ is the $G$-representation associated with output $\sR^{c_\text{out}}$.
For the \textit{scalar} value map, $\rho_\text{out}$ is identity, or trivial representation.

%They satisfy following equivariance conditions:
%\begin{align}
%%f_\theta &: \gS \times \gA \to \gS :& \rho_\gS(g) \cdot f_\theta(\vs_{t}, \va_{t}) &=  f_\theta(\rho_\gS(g) \cdot \vs_{t}, \rho_\gA(g) \cdot \va_{t}) \\
%%P_\theta &: \gS \times \gA \times \gS \to \sR^+  :& P_\theta(\vs_{t+1} \mid \vs_{t}, \va_{t}) &=  P_\theta(\rho_\gS(g) \cdot \vs_{t+1} \mid \rho_\gS(g) \cdot \vs_{t}, \rho_\gA(g) \cdot \va_{t}) \\
%R_\theta &: \gS \times \gA \to \sR :& {R}_\theta(\vs_{t}, \va_{t} ) &=  {R}_\theta(\rho_\gS(g) \cdot \vs_{t}, \rho_\gA(g) \cdot \va_{t}) \\
%Q_\theta &: \gS \times \gA \to \sR :& {Q}_\theta(\vs_{t}, \va_{t} ) &=  {Q}_\theta(\rho_\gS(g) \cdot \vs_{t}, \rho_\gA(g) \cdot \va_{t}) \\
%V_\theta &: \gS \to \sR :& {V}_\theta(\vs_{t}) &=  {V}_\theta(\rho_\gS(g) \cdot \vs_{t})
%\end{align}

For any group element $g \in \mathrm{E}(d) = \sR^d \rtimes \mathrm{O}(d)$, we transform the Bellman (optimality) operator step-by-step and show that it is equivariant under $\mathrm{E}(d)$:
\begin{align}
L_g \left[ \gT [V] \right](\vs) 
& \stackrel{\text{(1)}}{=} \gT [V](g^{-1}\vs) \\
& \stackrel{\text{(2)}}{=} \max_{\va} R(g^{-1}\vs, \va) + \int d\vs' \cdot P(\vs' \mid g^{-1} \vs, \va) V(\vs') \\
& \stackrel{\text{(3)}}{=} \max_{\bar \va} R(g^{-1}\vs, g^{-1} \bar \va) + \int d(g^{-1} \bar{\vs}) \cdot P(g^{-1} \bar{\vs} \mid g^{-1} \vs, g^{-1}\va) V(g^{-1}  \bar{\vs}) \\
& \stackrel{\text{(4)}}{=} \max_{\bar \va} R(\vs, \bar \va) + \int d(g^{-1} \bar{\vs}) \cdot P(\bar{\vs} \mid \vs, \va) V(g^{-1} \bar{\vs}) \\
& \stackrel{\text{(5)}}{=} \max_{\bar \va} R(\vs, \bar \va) + \int d\bar{\vs} \cdot P(\bar{\vs} \mid \vs, \va) V(g^{-1} \bar{\vs}) \\
& \stackrel{\text{(6)}}{=} \gT [ L_g [V]](\vs) 
\end{align}

For each step:
\begin{itemize}
\item (1) By definition of the (left) group action on the feature map $V: \gS \to \sR$, such that $g\cdot V(\vs) = \rho_0(g) V(g^{-1} \vs) = V(g^{-1} \vs)$. Because $V$ is a scalar feature map, the output transforms under trivial representation $\rho_0(g) = \mathrm{Id}$.
\item (2) Substitute in the definition of Bellman operator.
\item (3) Substitute $\va = g^{-1} (g \va) = g^{-1} \bar \va$. Also, substitute $g^{-1} \bar \vs = \vs'$.
\item (4) Use the symmetry properties of Geometric MDP: $P(\vs' \mid \vs, \va) = P(g \cdot \vs \mid g \cdot \vs, g \cdot \va)$ and $R (\vs, \va) = R(g \cdot \vs, g \cdot \va)$.
\item (5) Because $g \in \mathrm{E}(d)$ is isometric transformations (translations $\sR^d$, rotations and reflections $\mathrm{O}(d)$) and the state space carries group action, the measure $ds$ is a $G$-invariant measure $d(gs) = ds$. Thus, $d \bar{\vs} = d(g^{-1} \bar{\vs})$. 
\item (6) By the definition of the group action on $V$.
\end{itemize}

The proof requires the MDP to be a Geometric MDP with Euclidean symmetry and the state space carries a group action of Euclidean group.
Therefore, the Bellman operator of a Geometric MDP is $\mathrm{E}(d)$-equivariant.
Additionally, we can also parameterize the dynamics and reward functions with neural networks, and the learned Bellman operator is also equivariant.

The proof is analogous to the case in \citep{zhao_integrating_2022}, where the symmetry group is $p4m=\sZ^2 \rtimes D_4$, which is a discretized subgroup of $\mathrm{E}(2)$.
A similar statement can also be found in symmetric MDP \citep{zinkevich_symmetry_2001}, MDP homomorphism induced from symmetry group \citep{ravindran2004algebraic}, and later work on symmetry in deep RL \citep{van_der_pol_mdp_2020,wang_mathrmso2-equivariant_2021}.

% \begin{theorem}
% % For a geometric MDP, value iteration is an $\mathrm{E}(d)$-equivariant geometric message passing.
% For a GMDP, value iteration is an $\mathrm{E}(d)$-equivariant geometric message passing.
% \end{theorem}

\todonote{just theorem 2 here}
We additionally discuss another theorem.

\textbf{\textit{Theorem 2}}
\textit{For a GMDP, value iteration is an $\mathrm{E}(d)$-equivariant geometric message passing.}

\textit{Proof.}
We prove by constructing value iteration with 
For a more rigorous account on the relationship between dynamic programming (DP) and message passing on \textit{non-geometric} MDPs, see \citep{dudzik_graph_2022}.

%The key step is to relax the requirement of using \textit{scalar} transition probability function $P$ and \textit{scalar} reward function $R$:
%The key step is to relax the requirement on \textit{scalar} reward function $R_\theta$ and \textit{scalar} value function $V_\theta$ and $Q_\theta$. 
%The original scalar version is given by:
Notice that they satisfy the following equivariance conditions:
\begin{align}
%f_\theta &: \gS \times \gA \to \gS :& \rho_\gS(g) \cdot f_\theta(\vs_{t}, \va_{t}) &=  f_\theta(\rho_\gS(g) \cdot \vs_{t}, \rho_\gA(g) \cdot \va_{t}) \\
P_\theta &: \gS \times \gA \times \gS \to \sR^+  :& P_\theta(\vs_{t+1} \mid \vs_{t}, \va_{t}) &=  P_\theta(\rho_\gS(g) \cdot \vs_{t+1} \mid \rho_\gS(g) \cdot \vs_{t}, \rho_\gA(g) \cdot \va_{t}) \\
R_\theta &: \gS \times \gA \to \sR :& {R}_\theta(\vs_{t}, \va_{t} ) &=  {R}_\theta(\rho_\gS(g) \cdot \vs_{t}, \rho_\gA(g) \cdot \va_{t}) \\
Q_\theta &: \gS \times \gA \to \sR :& {Q}_\theta(\vs_{t}, \va_{t} ) &=  {Q}_\theta(\rho_\gS(g) \cdot \vs_{t}, \rho_\gA(g) \cdot \va_{t}) \\
V_\theta &: \gS \to \sR :& {V}_\theta(\vs_{t}) &=  {V}_\theta(\rho_\gS(g) \cdot \vs_{t}) \\
\end{align}

We construct geometric message passing such that it uses \textit{scalar} messages and features and resembles value iteration.

Then, we can use geometric message passing network to construct value iteration, which is to iteratively apply Bellman operators.
We adopt the definition of geometric message passing based on \citep{brandstetter_geometric_2021} as follows.
\begin{align}
%\tilde{\mathbf{m}}_{i j} & =\phi_m\left(\tilde{\mathbf{f}}_i, \tilde{\mathbf{f}}_j,\left\|\mathbf{x}_j-\mathbf{x}_i\right\|^2, \tilde{\mathbf{a}}_{i j}\right) \\
\tilde{\mathbf{m}}_{i j} & =\phi_m\left(\tilde{\mathbf{f}}_i, \tilde{\mathbf{f}}_j, \tilde{\mathbf{a}}_{i j}\right) \\
\tilde{\mathbf{f}}_i^{\prime} & =\phi_f\left(\tilde{\mathbf{f}}_i, \sum_{j \in \mathcal{N}(i)} \tilde{\mathbf{m}}_{i j}, \tilde{\mathbf{a}}_i\right) .
\end{align}
%\todo{finish}
The tilde means they are steerable under $G$ transformations.

We want to construct value iteration:
\begin{align}
	Q(s,a) &= R(s,a) + \gamma \sum_{s'} P(s'|s, a) V(s') \\
	V'(s) &= \sum_a \pi(a|s) Q(s,a)
\end{align}

To construct a \textit{geometric graph}, we let vertices $\gV$ be states $\vs$ and edges $\gE$ be state-action transition $(\vs, \va)$ labelled by $\va$.
For the geometric features on the graph, there are node features and edge features.
Node features include maps/functions on the state space: $\gS \to \sR^D$, and edge features include functions on the state-action space $\gS \times \gA \to \sR^D$.

%For example, state value function $V: \gS \to \sR^C$ is node feature, and $Q$-value function $Q_\theta: \gS \times \gA \to \sR^C$ and reward function $R_\theta : \gS \times \gA \to \sR^{C}$ are edge features.
For example, state value function $V: \gS \to \sR$ is (scalar) node feature, and $Q$-value function $Q_\theta: \gS \times \gA \to \sR$ and reward function $R_\theta : \gS \times \gA \to \sR$ are edge features.
The message $\tilde{\mathbf{m}}_{i j}$ is thus a scalar for every edge: $\tilde{\mathbf{m}}_{i j} = \pi(a|s) Q(s,a)$, and $\tilde{\mathbf{f}}_i^{\prime}$ is updated value function $\tilde{\mathbf{f}}_i^{\prime} = V'(s)$.
It is possible to extend value iteration to vector form as in Symmetric Value Iteration Network and Theorem 5.2 in \citep{zhao_integrating_2022}, while we leave it for future work.

%%%%%%%%%%%%%%%%%%%%%%%%%%%%%%%%%%%%%%%%%%%%%%%%%%%%%%%%%%%%%%%%%%%%%%%%%%%%%%%%%%%%%%%%%%%%%%%%
% \subsection{Proofs: Theorem 3 and 4}

% NOTE: removed all theory on linearized case

%%%%%%%%%%%%%%%%%%%%%%%%%%%%%%%%%%%%%%%%%%%%%%%%%%%%%%%%%%%%%%%%%%%%%%%%%%%%%%%%%%%%%%%%%%%%%%%%
%%%%%%%%%%%%%%%%%%%%%%%%%%%%%%%%%%%%%%%%%%%%%%%%%%%%%%%%%%%%%%%%%%%%%%%%%%%%%%%%%%%%%%%%%%%%%%%%

\section{Algorithm Design of Equivariant TD-MPC}
\label{sec:add_algorithm}

We elaborate on the algorithm design in this section.

\paragraph{Invariance of \texttt{return}.}

We compute the expected return of sampled trajectories, and study how it is transformed: 
% Under expectation, the return is computed by the next equation.
% We can study how the return is transformed.
%
%given by $\texttt{return}(\tau_i) = \mathbb{E}_{\tau}\left[\gamma^H Q_\theta\left(\vs_H, \va_H\right)+\sum_{t=0}^{H-1} \gamma^t R_\theta\left(\vs_t, \va_t\right)\right]$
\begin{align}
%	\phi_{\Gamma} \triangleq \mathbb{E}_{\Gamma}\left[\gamma^H Q_\theta\left(\mathbf{z}_H, \mathbf{a}_H\right)+\sum_{t=0}^{H-1} \gamma^t R_\theta\left(\mathbf{z}_t, \mathbf{a}_t\right)\right]
	\texttt{return}({\tau}) & = \mathbb{E}_{\tau}\left[\gamma^H Q_\theta\left(\vs_H, \va_H\right)+\sum_{t=0}^{H-1} \gamma^t R_\theta\left(\vs_t, \va_t\right)\right] = \mathbb{E}_{\tau}\left[ U(\vs_{1:H}, \va_{1:H-1}) \right] \\
	\texttt{return}(g \cdot {\tau}) & = \mathbb{E}_{g \cdot \tau}\left[\gamma^H \rho_0(g) \cdot  Q_\theta\left(g \cdot \vs_H, g \cdot \va_H\right)+\sum_{t=0}^{H-1} \gamma^t \rho_0(g)\cdot  R_\theta\left(g \cdot \vs_t, g \cdot \va_t\right)\right] \label{eq:tranformed_traj} \\
	& = \int_{g\in G}\rho_0(g) dg \cdot  \mathbb{E}_{\tau}\left[ U(g \cdot \vs_{1:H}, g \cdot \va_{1:H-1})  \right] \label{eq:extracted_g} \\
		& = \mathbf{1} \cdot  \mathbb{E}_{\tau}\left[ U(\vs_{1:H}, \va_{1:H-1})  \right] = \texttt{return}({\tau}) \label{eq:g_simplify}
\end{align}

%Suppose the term is denoted as $U(\vs_{1:H}, \va_{1:H-1})$

In Equation~\ref{eq:tranformed_traj}, we use $\rho_0(g) = \mathbf  1$ to denote that the output is not transformed, so we may extract the term out.
In Equation~\ref{eq:extracted_g}, $dg$ is a Haar measure that absorbs the normalization factor, and we can extract the term from expectation.
Equation~\ref{eq:g_simplify} uses the invariance of $Q_\theta$ and $R_\theta$.
In other words, the return under the $G$-orbit of trajectories is the same, thus \texttt{return} is $G$-invariant.

\paragraph{Equivariance of $G$\texttt{-sample}.}

In Model Predictive Path Integral (MPPI)~\citep{williams_information_2017}, we sample $N$ actions from a random Gaussian distribution $\mathcal{N}(\mu, \sigma^2 I)$, denoted as $\sA = \{ a_i \}_{i=1}^N$. However, since $\mu$ and $\sigma$ are not state-dependent, CEM/MPPI does not satisfy the condition of equivariance, which requires that rotating the input results in a rotated output.
To address this, we propose a solution - augmenting the action sampling by transforming with all elements in the group $G$: $G\sA =\{ g \cdot a_i \mid g \in G \}_{i=1}^N$. This approach ensures that our method can handle different orientations and maintain the property of equivariance.

To validate our approach, we first demonstrate the equivariance condition mathematically. 
We assume that $(s_0, a_0)$ gives the maximum value 
$$a_0 = \arg\max_{a\in G\sA} Q(s_0, a).$$
If we consider 
$$g \cdot a_0 = g \cdot \arg\max_{a\in G\sA} Q(s_0, a) = \arg\max_{a\in G\sA} Q(g \cdot s_0, a),$$ 
it implies that if we rotate the state to $g\cdot s_0$, we expect $g\cdot a_0$ to still provide the maximum $Q$-value so that $\arg\max$ can select it. 
The proof is validated using the invariance of $Q$, $Q(g \cdot s, g \cdot a) = Q(s,a)$.
Hence,
$$a_0' = \arg\max_{a\in G\sA} Q(g \cdot s_0, a) = \arg\max_{a\in G\sA} Q(s_0, g^{-1} \cdot a).$$
By comparing these two equations, we find that $a_0' = g \cdot a_0$.

Note that when not augmenting $\sA$, it is not guaranteed that $g \cdot a_0$ exists in $\sA$.
However, when the number of samples approaches infinity, $g \cdot a_0$ can get close to some element in $\sA$.

The proof can be directly applied to multiple steps, as $\texttt{return}$ is also $G$-invariant.

\todoil{
We then empirically measure the equivariance error under finite samples for three different methods: (1) the non-equivariant method, (2) the equivariant model only, and (3) the equivariant model with G-augmented sampling.
}

\todolsw{(previous; hope my version addressed) I did not get a sense of what was done to fix the problem.}

%%%%%%%%%%%%%%%%%%%%%%%%%%%%%%%%%%%%%%%%%%%%%%%%%%%%%%%%%%%%%%%%%%%%%%%%%%%%%%%%%%%%%%%%%%%%%%%%
%%%%%%%%%%%%%%%%%%%%%%%%%%%%%%%%%%%%%%%%%%%%%%%%%%%%%%%%%%%%%%%%%%%%%%%%%%%%%%%%%%%%%%%%%%%%%%%%

\section{Implementation Details and Additional Evaluation}
\label{sec:add_results}

%%%%%%%%%%%%%%%%%%%%%%%%%%%%%%%%%%%%%%%%%%%%%%%%%%%%%%%%%%%%%%%%%%%%%%%%%%%%%%%%%%%%%%%%%%%%%%%%
\subsection{Implementation Details: Equivariant TD-MPC}

We mostly follow the implementation of TD-MPC \citep{hansen_temporal_2022}.
The training of TD-MPC is end-to-end, i.e., it produces trajectories with a learned dynamics and reward model and predicts the values and optimal actions for those states.
It closely resembles MuZero \citep{schrittwieser_mastering_2019} while uses MPPI (Model Predictive Path Integral \citep{williams_model_2015,williams_information_2017}) for continuous actions instead of MCTS (Monte-Carlo tree search) for discrete actions.
It inherits the drawbacks from MuZero - the dynamics model is trained only from reward signals and may collapse or experience instability on sparse-reward tasks.
This is also the case for the tasks we use: \texttt{PointMass} and \texttt{Reacher} and their variants, where the objectives are to reach a goal position.
% We

%%%%%%%%%%%%%%%%%%%%%%%%%%%%%%%%%%%%%%%%%%%%%%%%%%%%%%%%%%%%%%%%%%%%%%%%%%%%%%%%%%%%%%%%%%%%%%%%

\subsection{Experimental Details}

We implement $G$-equivariant MLP using \texttt{escnn} \citep{weiler_general_2021} for policy, value, transition, and reward network, with 2D and 3D discrete groups.
For all MLPs, we use two layers with $512$ hidden units. The hidden dimension is set to be $48$ for non-equivariant version, and the equivariant version is to keep the same number of free parameters, or \texttt{sqrt} strategy.

For example, for $D_8$ group, \texttt{sqrt} strategy (to keep same free parameters) has number of hidden units divided by $\sqrt{|D_8|} = \sqrt{16}=4$.
The other strategy is to make equivariant networks' input and output be compatible with non-equivariant ones: \textit{\texttt{linear}} strategy, which keeps same input/output dimensions (number of hidden units divided by $|D_8| = 16$).

We use two strategies: \texttt{sqrt} strategy (to keep same free parameters, number of hidden units divided by $\sqrt{|D_8|} = \sqrt{16}=4$) on specifying the number of hidden units, we use \textit{\texttt{linear}} strategy that keeps same input/output dimensions (number of hidden units divided by $|D_8| = 16$)

The hidden space uses \textit{regular} representation, which is common for discrete equivariant network \citep{cohen_group_2016,weiler_general_2021,zhao_integrating_2022}.

%%%%%%%%%%%%%%%%%%%%%%%%%%%%%%%%%%%%%%%%%%%%%%%%%%%%%%%%%%%%%%%%%%%%%%%%%%%%%%%%%%%%%%%%%%%%%%%%

\subsection{Additional Results}
\label{subsec:add-results-appendix}

\begin{figure*}[t]
\centering
%\vspace{-20pt}
\subfigure{
\includegraphics[width=0.6\linewidth]{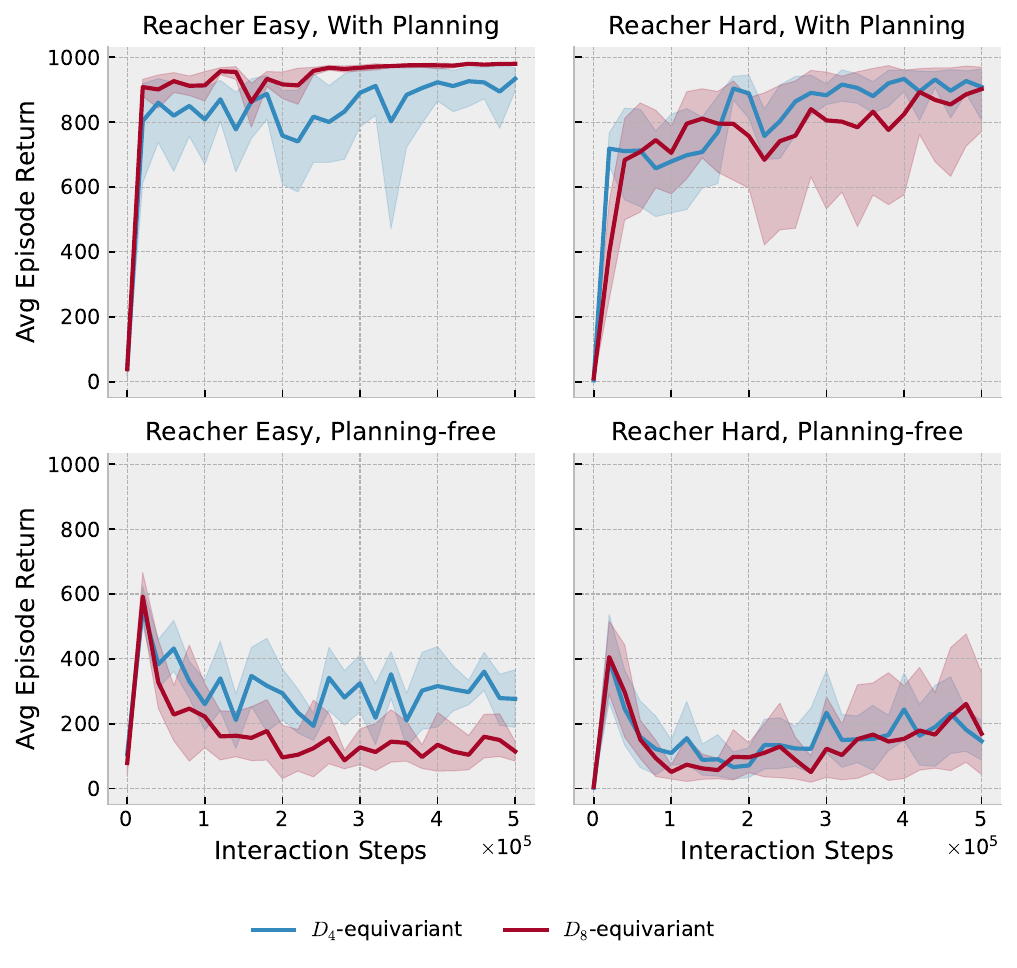}
}
%
% \vspace{-15pt}
\centering
\caption{
Ablation study on planning component.
}
% \vspace{-14pt}
\label{fig:ablation-planning-component}
\end{figure*}

\paragraph{Ablation on model-based vs. model-free (``planning-free'').}

We ablate the use of planning component in equivariant version of TD-MPC, which is to justify why we aim to build model-based version of equivariant RL algorithm over model-free counterparts.
The results are shown in Figure~\ref{fig:ablation-planning-component}. On both \texttt{Reacher} Easy and Hard, with planning, the performance is much better.

\begin{figure*}[t]
\centering
%\vspace{-20pt}
\subfigure{
\includegraphics[width=0.5\linewidth]{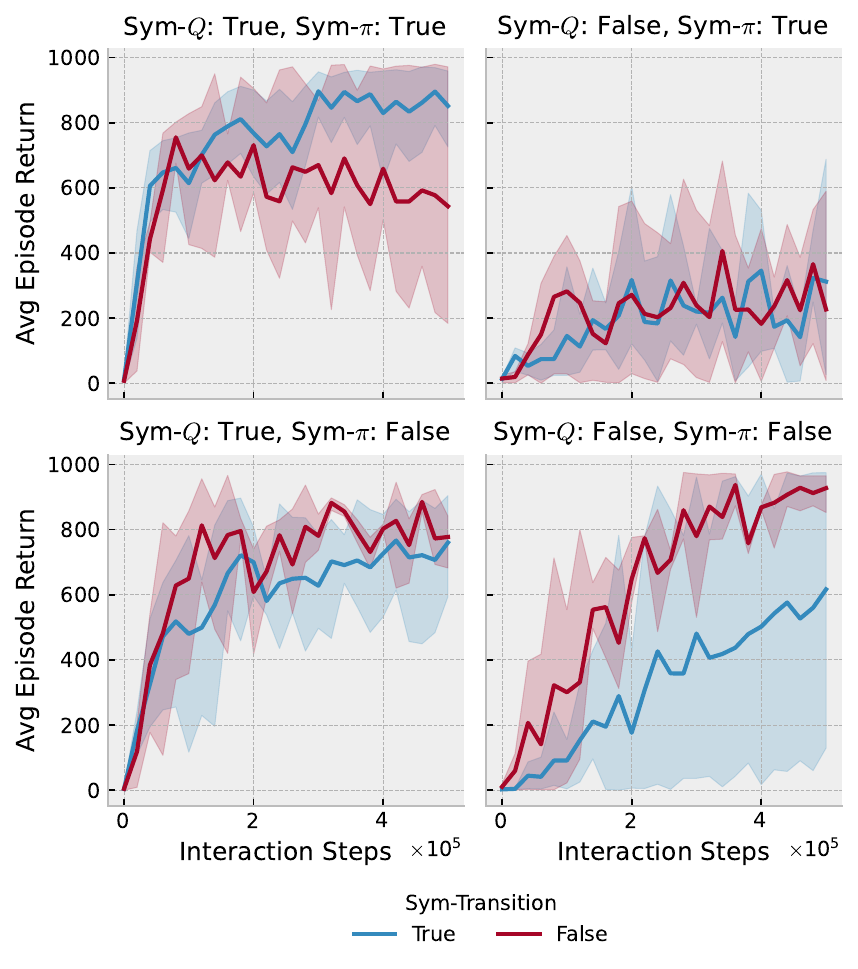}
}
%
% \vspace{-15pt}
\centering
\caption{
Ablation study on equivariant components, using \texttt{Reacher} Hard with $D_8$ symmetry group.
}
% \vspace{-14pt}
\label{fig:ablation-equiv-componets}
\end{figure*}

\paragraph{Ablation on equivariant components.}

Recall that we have several equivariant components in equivariant TD-MPC:
\begin{align}
f_\theta &: \gS \times \gA \to \gS :& \rho_\gS(g) \cdot f_\theta(\vs_{t}, \va_{t}) &=  f_\theta(\rho_\gS(g) \cdot \vs_{t}, \rho_\gA(g) \cdot \va_{t}) \\
R_\theta &: \gS \times \gA \to \sR :& {R}_\theta(\vs_{t}, \va_{t} ) &=  {R}_\theta(\rho_\gS(g) \cdot \vs_{t}, \rho_\gA(g) \cdot \va_{t}) \\
Q_\theta &: \gS \times \gA \to \sR :& {Q}_\theta(\vs_{t}, \va_{t} ) &=  {Q}_\theta(\rho_\gS(g) \cdot \vs_{t}, \rho_\gA(g) \cdot \va_{t}) \\
\pi_\theta &: \gS \to \gA  :& \rho_\gA(g) \cdot {\pi}_\theta(\cdot \mid \vs_{t}) &=  {\pi}_\theta(\cdot \mid \rho_\gS(g) \cdot \vs_{t})
\end{align}

We experiment to enable and disable each of them: (1) transition network: dynamics $f$ and reward $R$, (2) value network: $Q$, and (3) policy network $\pi$.
Note that to make equivariant and non-equivariant components compatible, we need to make sure the input and output dimensions match.

We show the results on \texttt{Reacher} Hard with $D_8$ symmetry group in Fig~\ref{fig:ablation-equiv-componets}.
Instead of using \texttt{sqrt} strategy (to keep same free parameters, number of hidden units divided by $\sqrt{|D_8|} = \sqrt{16}=4$) on specifying the number of hidden units, we use \textit{\texttt{linear}} strategy that keeps same input/output dimensions (number of hidden units divided by $|D_8| = 16$). Thus, the performance of fully non-equivariant model and fully equivariant model are not directly comparable, because the number of free parameters in fully equivariant one is much smaller.

The results show the relative importance of value, policy, and transition.
It shows the most important equivariant component is $Q$-value network. It is reasonable because it has been used intensively in predicting into the future, where generalization and training efficiency are very important and benefit from equivariance.

\begin{figure*}[t]
\centering
%\vspace{-20pt}
\subfigure{
\includegraphics[width=1.\linewidth]{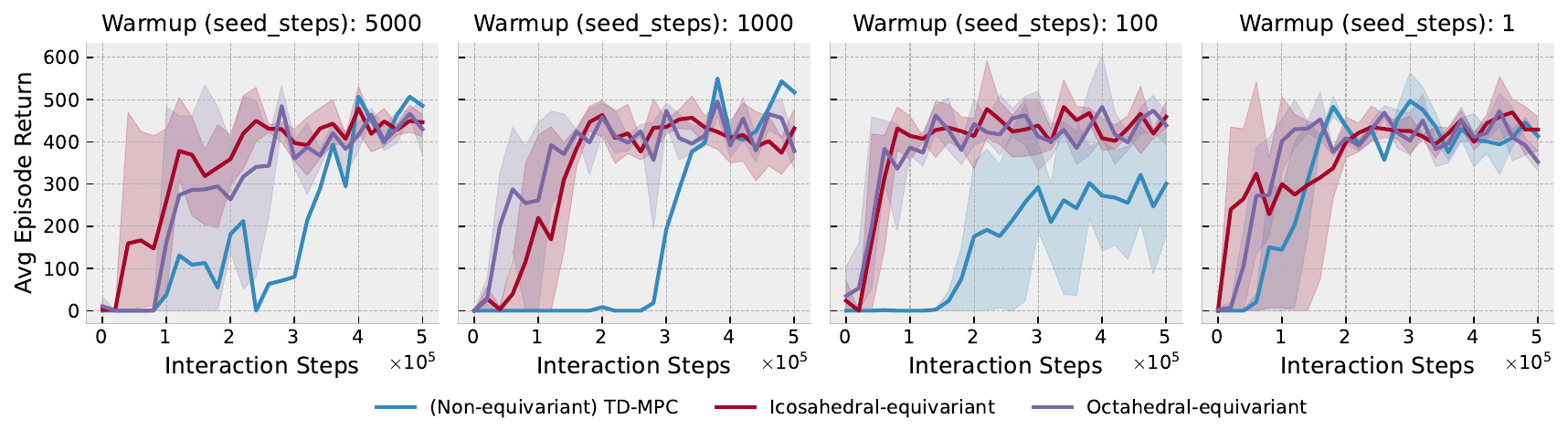}
}
%
% \vspace{-15pt}
\centering
\caption{
Ablation study on number of warmup episodes on \texttt{PointMass} \textit{3D with small target}.
}
% \vspace{-14pt}
\label{fig:ablation-warmup}
\end{figure*}

\paragraph{Hyperparameter of amount of warmup.}
We experiment different number of warmup episodes, called \texttt{seed steps} in TD-MPC hyperparameter.
We find this is a critical hyperparameter for (non-equivariant) TD-MPC.
One possible reason is that TD-MPC highly relies on joint training and may collapse when the transition model is stuck at some local minima. This warmup hyperparameter controls how many episodes TD-MPC collects before starting actual training.

We test using different numbers on \texttt{PointMass} \textit{3D with small target}.
The results are shown in Figure~\ref{fig:ablation-warmup}, which demonstrate that our equivariant version is robust under all choices of warmup episodes, even with little to none warmup.
The non-equivariant TD-MPC is very sensitive to the choice of warmup number.

\begin{figure*}[t]
\centering
%\vspace{-20pt}
\subfigure{
\includegraphics[width=0.5\linewidth]{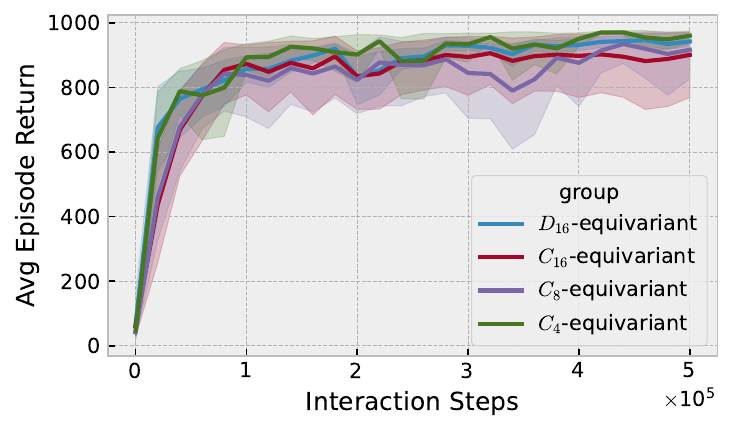}
}
%
% \vspace{-15pt}
\centering
\caption{
Ablation study on symmetry group on \texttt{Reacher} Hard.
}
% \vspace{-14pt}
\label{fig:ablation-symmetry}
\end{figure*}

\paragraph{Ablation on symmetry groups.}

We also do ablation study on the choice of discrete subgroups.
We run experiments on \texttt{Reacher} Hard to compare 2D discrete rotation/dihedral groups: $C_4, C_8, C_{16}, D_{16}$, using $1$ warmup episode.

The results are shown in Fig~\ref{fig:ablation-symmetry}. We find using groups larger than $C_8$ does not bring additional improvement on this specific task, \texttt{Reacher} Hard.
In the main paper, we thus use $D_4, D_8$ to balance the performance and computation time and memory use.

\todoilzlf{update this figure - using latest groups}

\paragraph{Comparing reference frames and state features.}

This experiment studies the balance between reference frames and the choice of state features.
In the theory section, we emphasize that kinematic constraints introduce local reference frames.

Here, we study a specific example: \texttt{Reacher} (Easy and Hard).
The second joint has angle $\theta_2$ and angular velocity $\dot \theta_2$ relative to the first link.

For \textit{local} reference frame version, we use
\begin{equation}
    \left(\theta_1, \theta_2, \dot{\theta}_1, \dot{\theta}_2, x_g-x_f, y_g-y_f\right) \Rightarrow\left(\cos \theta_1, \sin \theta_1, \cos \theta_2, \sin \theta_2, \dot{\theta}_1, \dot{\theta}_2, x_g-x_f, y_g-y_f\right)
\end{equation}
Thus, $\cos \theta_1, \sin \theta_1$ is transformed under standard representation $\rho_1$ and $\cos \theta_2, \sin \theta_2$ is transformed under trivial representation $\rho_0 \oplus \rho_0$.

For the \textit{global} reference frame version, we compute the global location of the end-effector (tip) by adding the location of the first joint.
Thus, the global position is transformed also under standard representation now $\rho_1$.

\begin{figure*}[t]
\centering
%\vspace{-20pt}
\includegraphics[width=0.8\linewidth]{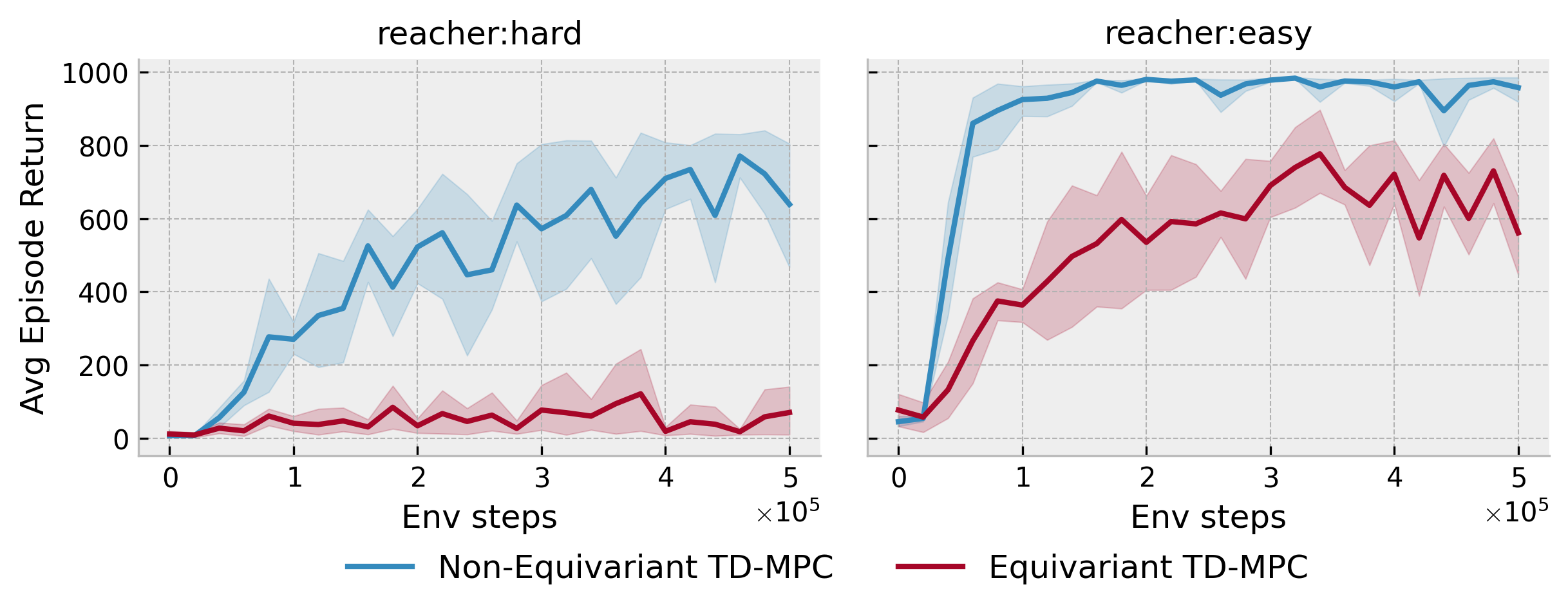}
% \vspace{-15pt}
\centering
\caption{
Results for global reference frame on \texttt{Reacher}.
}
% \vspace{-14pt}
\label{fig:env-reward-curves-global}
\end{figure*}

We show the results in Figure~\ref{fig:env-reward-curves-global}.
Evaluation reward curves for non-equivariant and equivariant TD-MPC over $5$ runs using global frames. Error bars denote $95\%$ confidence intervals. Non-equivariant TD-MPC outperforms equivariant TD-MPC.
Surprisingly, we find using global reference frame where the second joint is associated with standard representation (equivariant feature, instead of invariant feature) brings much worse results, compared to the local frame version in the main paper.
One possibility is that it is more important to encode kinematic constraints (e.g., the length of the second link is preserved in $\cos \theta_2, \sin \theta_2$), compared to using equivariant feature.

%%%%%%%%%%%%%%%%%%%%%%%%%%%%%%%%%%%%
% \begin{figure*}[t]
% \centering
% %\vspace{-20pt}
% \subfigure{
% \includegraphics[width=0.48\linewidth]{results/reacher_easy_local.png}
% }
% %
% \subfigure{
% \includegraphics[width=0.48\linewidth]{results/reacher_hard_local.png}
% }
% % \vspace{-15pt}
% \centering
% \caption{
% Evaluation reward curves for non-equivariant and equivariant TD-MPC over $5$ runs. Error bars denote $95\%$ confidence intervals. Equivariant TD-MPC is more sample efficient than non-equivariant TD-MPC.
% }
% % \vspace{-14pt}
% \label{fig:env-reward-curves}
% \end{figure*}

%%%%%%%%%%%%%%%%%%%%%%%%%%%%%%%%%%%%

%%%%%%%%%%%%%%%%%%%%%%%%%%%%%%%%%%%%%%%%%%%%%%%%%%%%%%%%%%%%%%%%%%%%%%%%%%%%%%%%%%%%%%%%%%%%%%%%
%%%%%%%%%%%%%%%%%%%%%%%%%%%%%%%%%%%%%%%%%%%%%%%%%%%%%%%%%%%%%%%%%%%%%%%%%%%%%%%%%%%%%%%%%%%%%%%%

% NOTE: removed other additional sections, e.g., theory and more
% See /unused/math_background.tex

\end{document}